\documentclass[journal]{IEEEtran}
\ifdefined\MARKEDVERSION
  \usepackage[normalem]{ulem}
  \usepackage[
    markup=underlined,
    commandnameprefix=ifneeded,
    defaultcolor=blue
  ]{changes}
  \setdeletedmarkup{\textcolor{red}{\scriptsize\sout{#1}}}
\else
  \usepackage[
    final,
    commandnameprefix=ifneeded
  ]{changes}
\fi
\newcommand{\revisedblock}{\ifdefined\MARKEDVERSION\color{blue}\fi}
\usepackage{microtype}
\usepackage[nolist]{acronym}
\usepackage{orcidlink}
\usepackage{xr-hyper}
\usepackage{hyperref}
\hypersetup{colorlinks=true, allcolors=blue}
\acrodef{IEEE}{Institute of Electrical and Electronics Engineers}
\acrodef{LTT}{learn then test}
\acrodef{ZS}{zero-shot}
\acrodef{TL}{transfer learning}
\acrodef{TD}{target-domain}
\acrodef{RA}{risk-averse}
\acrodef{EO}{efficiency-oriented}
\acrodef{XGB}{extreme gradient boosting}
\acrodef{GC}{global conformal}
\acrodef{MU}{model uncertainty}
\acrodef{CP}{conformal prediction}
\acrodef{BV}{band validity}
\acrodef{LP}{lightpath}
\acrodef{OSNR}{optical signal-to-noise ratio}
\acrodef{ZSM}{zero-touch network management}
\acrodef{LAM}{large AI model}
\acrodef{SLA}{service level agreement}
\acrodef{QES}{quality efficiency score}
\acrodef{CQI}{Composite Quality Index}
\acrodef{CoT}{Chain-of-Thought}
\acrodef{RL}{reinforcement learning}
\acrodef{AI}{artificial intelligence}
\acrodef{AI/ML}{artificial intelligence \& machine learning}
\acrodef{ANN}{artificial neural network}
\acrodef{ASE}{amplified spontaneous emissions}
\acrodef{CUT}{channel under test}
\acrodef{DRL}{deep reinforcement learning}
\acrodef{EGN}{enhanced Gaussian noise}
\acrodef{EON}{elastic optical network}
\acrodef{GT}{ground truth}
\acrodef{GSNR}{generalized signal-to-noise ratio}
\acrodef{KDE}{kernel density estimation}
\acrodef{LLM}{large language model}
\acrodef{LP}{lightpath}
\acrodef{MAE}{mean absolute error}
\acrodef{MF}{modulation format}
\acrodef{ML}{machine learning}
\acrodef{NLI}{non-linear impairments}
\acrodef{PDF}{probability density function}
\acrodef{QoT}{quality of transmission}
\acrodef{RMSA}{routing, modulation and spectrum assignment}
\acrodef{RMSE}{root mean squared error}
\acrodef{XGB}{XGBoost}
\acrodef{IBN}{intent-based networking}
\acrodef{BER}{bit error rate}
\acrodef{XAI}{explainable artificial intelligence}
\acrodef{ADON}{autonomous driving optical network}
\acrodef{SHAP}{SHapley Additive exPlanations}
\acrodef{Bu-SMF}{bundled single-mode fiber}
\acrodef{WC-MCF}{weakly coupled multicore fiber}
\acrodef{FrJ-Sw}{fractional-joint switching}
\acrodef{J-Sw}{joint switching}
\acrodef{Ind-Sw}{independent switching}
\acrodef{SDM}{space-division multiplexing}
\acrodef{RA}{risk-averse}
\acrodef{EO}{efficiency-oriented}
\acrodef{CRC}{conformal risk control}
\acrodef{QR}{quantile regression}
\acrodef{EDFA}{erbium-doped fiber amplifier}
\acrodef{FAS}{fiber-amplifier-switch}
\acrodef{FA}{fiber-amplifier}
\acrodef{FNR}{false negative rate}
\acrodef{FPR}{false positive rate}

\usepackage{pgfplots}
\usepackage{tikz}
\usepackage{xcolor}
\usetikzlibrary{patterns}
\pgfplotsset{compat=1.18}

\usepackage{subcaption}
\usepackage{adjustbox}

\definecolor{BaselineBlue}{HTML}{4C78A8}
\definecolor{RiskTeal}{HTML}{72B7B2}
\definecolor{EfficiencyOrange}{HTML}{DD8452}
\definecolor{LttBlue}{HTML}{0072B2}
\definecolor{OrangeBand}{HTML}{E69F00}
\definecolor{TruthRed}{HTML}{D55E00}
\definecolor{PredBlack}{HTML}{333333}

\usepackage{amsmath}
\usepackage{amssymb}
\usepackage{booktabs}
\ifCLASSOPTIONcompsoc
 \usepackage{subcaption}
\else

\fi
\usepackage{svg}
\begin{document}
\title{Conformal Risk Control for QoT-Aware Lightpath Feasibility Under Technology-Driven Domain Shift
% \thanks{Received XX XXX 2026; revised XX XXX 2026; accepted XX XXX 2026. Date of publication XX XXXX 2026; date of current version XX XXX 2026.}
\thanks{This work was supported by the EUREKA Celtic-Next Flagship SUSTAINET-Advance project, funded by VINNOVA (2025-02987, Sweden) and Innosuisse (No. 119.588 INT-ICT, Switzerland). (Corresponding author: Kiarash Rezaei.)}\thanks{Kiarash Rezaei, Paolo Monti, and Carlos Natalino are with the Department of Electrical Engineering, Chalmers University of Technology, 412 96 Gothenburg, Sweden. (e-mail: kiarashr@chalmers.se, mpaolo@chalmers.se, carlos.natalino@chalmers.se).}\thanks{Omran Ayoub is with the University of Applied Sciences and Arts of Southern Switzerland, 6928 Lugano, Switzerland. (e-mail: omran.ayoub@supsi.ch).}\thanks{Color versions of one or more figures in this article are available at
\url{https://doi.org/10.1109/JLT.2026.3752946}.}
\thanks{
Digital Object Identifier \href{https://doi.org/10.1109/JLT.2026.3752946}{10.1109/JLT.2026.3752946}
}
}
\author{
    \IEEEauthorblockN{Kiarash Rezaei\IEEEauthorrefmark{1}\orcidlink{0009-0003-6166-2614}, 
    Omran Ayoub\IEEEauthorrefmark{2}\orcidlink{0000-0002-3884-3594},
    Paolo Monti\IEEEauthorrefmark{1}\orcidlink{0000-0002-5636-9910}, Carlos Natalino\IEEEauthorrefmark{1}\orcidlink{0000-0001-7501-5547}}
    }
\markboth{Journal of Lightwave Technology}% , VOL. XX, NO. XX, XXX, 2026.
{Rezaei \MakeLowercase{\textit{et al.}}: Conformal Risk Control}
\maketitle
\begin{abstract}
\Ac{AI/ML}-based regression models are increasingly used to support operational decision-making in optical networks, but they typically expose only point predictions, which carry no statistically controlled information about their own reliability.
This paper proposes a post-hoc \ac{CRC} framework that \replaced{keeps a pretrained \ac{AI/ML} regressor fixed, provides a finite-sample guarantee on uncertainty band miscoverage, and maps the band to operational decisions through policy rules. The guarantee applies to interval miscoverage, while the resulting decision-error rates are evaluated empirically.}{transforms a pretrained \ac{AI/ML} regression model into a risk-controlled decision-support mechanism. The framework keeps the predictor fixed, constructs uncertainty bands using calibration data, and maps these bands into operational decisions through policy rules.}
We instantiate the framework for \ac{QoT}-aware \ac{LP} feasibility assessment, in which a regressor predicts \ac{OSNR} under domain shifts caused by upgrades in fiber type, amplifier technology, and switching architecture.
Feature shift contribution scores identify the \ac{LP} features whose relation to \ac{QoT} changes the most across a technology transition, and the resulting bands are translated into feasibility decisions by two policies, allowing operators to tune the trade-off between conservatively rejecting or aggressively accepting \ac{LP} realizations.
Experiments on open \ac{SDM} \ac{QoT} datasets over the CONUS topology show that, for a target coverage of 90\%, \replaced{the uncertainty bands meet the target in all 170 trials spanning 17 domain shifts and ten calibration--test partitions, whereas a source-only \ac{QR} baseline meets it in none.}{the uncertainty bands achieve above-90\% empirical coverage of the true \ac{OSNR} values across all evaluated domain shift families.}
Under severe domain shifts, the \ac{EO} policy empirically improves \added{\ac{ZS}}\deleted{zero-shot} feasibility-decision accuracy from about 85\% to above 99\%, approaching a \ac{TL} policy while avoiding retraining.
These results show that \replaced{the framework provides finite-sample control of uncertainty band miscoverage for regression-based optical network decision support, with potential applications beyond \ac{QoT}-aware \ac{LP} feasibility assessment.}{the proposed framework supports statistically controlled decision-making for regression-based optical network applications, with potential applications beyond \ac{QoT}-aware \ac{LP} feasibility assessment.}
\end{abstract}

\IEEEpeerreviewmaketitle
\ifdefined\MARKEDVERSION
  % The marked copy carries both old and new wording; use the IEEE small text
  % size for its body so the revision copy remains within the page limit.
  \small
\fi
\acresetall

\section{Introduction}
\label{sec:intro}

% Paragraph 1: Optical network motivation
Optical transport networks are evolving toward increasingly agile and heterogeneous infrastructures, driven by coherent transceivers, flexible-grid elastic optical networking, and emerging \ac{SDM} technologies \cite{Casellas:25,Natalino_2025_ICMLCN}.
These advances expand the control-plane design space by introducing degrees of freedom in routing, modulation-format selection, and spectrum/spatial/waveband allocation, but they also intensify the need for fast and reliable physical-layer feasibility assessment \cite{Lohani_2026_JOCN}.
In particular, \ac{QoT} estimation for \emph{unestablished} \acp{LP} is a key enabler of impairment-aware provisioning, reconfiguration, and monitoring in dynamic optical networks \cite{KLINKOWSKI201858, Yang:18}, and \ac{AI/ML} regressors can approximate analytical \ac{QoT} models with high accuracy when detailed physical-layer modeling is too expensive or impractical.

Operational deployment, however, demands more than the strong average accuracy typically observed during model development \cite{Natalino_2026_EngineeringTrustAIDriven}. A \emph{trustworthy} model must also remain reliable when deployment conditions depart from the training data, and expose how much confidence each individual prediction deserves.
Network evolution stresses both requirements at once.
Differences between training and operational conditions arise from unintentional processes, such as component aging, and from intentional interventions, such as technology upgrades \cite{Lechowicz_2025_QoTEstimationMarginDriven}.
The latter are usually the more demanding case, because an upgrade in fiber, amplifier, or switching technology modifies not only the distribution of the \ac{LP} descriptors, but also their underlying relationship with \ac{QoT}.
Such a source--target domain shift silently erodes the accuracy of a regressor trained on the pre-upgrade infrastructure, and a point prediction gives no signal that it has occurred.

Existing work \replaced{has studied accuracy loss under domain shift and the reliability of individual predictions}{addresses these issues} separately, and neither line of work assesses their impact on operational decisions.
Domain shift mitigation methods recover target-domain accuracy, \replaced{through}{but require} retraining or adapting the model~\cite{Gu:23,Zhou:24,Cai:24,Wang:24}, \replaced{but require the ability to update the predictor and sufficient}{which presumes access to the model internals and enough} target-domain data to update it.
Uncertainty-aware methods, in turn, are frequently model-specific or heuristic~\cite{Tanimura:19, nicola-22, nicola-23, somaiieh:23,Guo:26}, and report a confidence value rather than committing to an operational choice.
Complementing them, \ac{CP} equips black-box \ac{AI/ML} models with distribution-free uncertainty estimates under mild assumptions~\cite{angelopoulos2022gentleintroductionconformalprediction}, applied post-hoc using \deleted{a small }held-out \added{labeled }target-domain dataset and without retraining, while its connection to \ac{CRC}~\cite{Angelopoulos:22CRC} extends conformal methods from coverage guarantees to broader, user-defined risk-control objectives.
\replaced{However, to the best of our knowledge, no prior work in the literature translates a risk-controlled uncertainty band into optical network decisions,}{ conformal uncertainty into risk-controlled optical network decisions,}
precisely when a trained \ac{AI/ML} model is reused after fiber, amplifier, or switching upgrades.

This paper closes that gap \replaced{by proposing}{with} a post-hoc \ac{CRC} decision-support framework that leaves the deployed regressor untouched, organized in three layers:
\emph{(i)} a fixed \ac{AI/ML} regressor,
\emph{(ii)} a \ac{CRC} layer that converts each point prediction into a risk-controlled uncertainty band from a small target-domain calibration set,
and \emph{(iii)} a policy layer that maps the prediction and band into an operational decision.
We instantiate \replaced{the framework}{it} for \ac{QoT}-aware \ac{LP} feasibility assessment under technology-driven domain shift, building on our preliminary conference work~\cite{kiarash2026ofc}, which paired split-conformal bands with feasibility policies and validated the idea on a single source--target transition, and advancing it along four directions.

First, we strengthen the statistical foundations of the conformal band.
Whereas~\cite{kiarash2026ofc} sets its width to an empirical quantile of the calibration residuals, which controls coverage on average over calibration sets, in this work we select it through an \ac{LTT}-based \ac{CRC} procedure~\cite{angelopoulos2022learntestcalibratingpredictive} from an upper confidence bound on the miscoverage risk that holds uniformly over a grid of candidate widths.
This yields the guarantee that, even with the specific and possibly small calibration set available, the selected band keeps the probability that the true \ac{QoT} value falls outside the band below $\alpha$, with confidence of at least $1-\delta$.
The statement remains valid even though the width is selected using that same calibration set.
Second, we broaden the evaluation to 17 source--target scenarios grouped into six shift families, covering fiber, amplifier, and switching upgrades in isolation and in combination.
While the transition studied in~\cite{kiarash2026ofc} changes fiber and amplifier jointly, our isolated evaluation results identify the amplifier as the dominant contributor.
Third, we characterize each shift using feature shift contribution scores that identify the \ac{LP} features whose relation to \ac{QoT} changes the most, and we delimit what such feature-level scores can and cannot reveal about the impact on the deployed model.
\replaced{Fourth, we map the risk-controlled band to feasibility decisions through two policies, \ac{RA}, which prioritizes avoiding unsafe acceptance, and \ac{EO}, which prioritizes avoiding overly conservative rejection. Their decision-error rates are assessed empirically.}{Fourth, we map the risk-controlled band to feasibility decisions through two policies, \ac{RA}, which avoids accepting infeasible \acp{LP}, and \ac{EO}, which reduces overly conservative rejection.}

We evaluate the framework on open \ac{SDM} \ac{QoT} datasets~\cite{Akbari:25} over the CONUS topology, \replaced{comparing it with a \ac{ZS} source model, target-trained \ac{TL}, and a source-only \ac{QR} uncertainty baseline. Across ten calibration--test partitions of each of the 17 scenarios, the \ac{LTT} bands meet the empirical coverage target in all 170 trials.}{ comparing the proposed policies with two point-prediction baseline policies: a \ac{ZS} policy that applies the fixed source-trained regressor directly to the target domain, and a \ac{TL} policy that adapts the regressor using target-domain data and serves as a practical upper reference. The \ac{LTT} bands maintain above-target empirical coverage across all shift families.}
The \ac{EO} policy improves accuracy under the most challenging shifts from about 85\% for the \ac{ZS} policy to above 99\%, approaching the \ac{TL} policy without retraining.
The \ac{RA} policy provides a conservative operating mode with near-zero \ac{FPR}.
These results indicate that the proposed approach can turn source-trained \ac{QoT} regressors into practical decision-support tools under technology-driven deployment shifts, and suggest that the same post-hoc risk-control principle may also apply to other regression-driven optical network tasks.

\section{Related Work}
\label{sec:related_work}

\ac{AI/ML}-based \ac{QoT} estimation has been widely studied in optical networks.
Early studies formulated the problem as feasibility classification \cite{Panayiotou:17, Rottondi:18}, while later work considered regression-based \ac{QoT} estimation \cite{Ibrahimi:21}.
Surveys and tutorials have \deleted{further} consolidated this research direction by systematizing model families, feature engineering practices, and deployment considerations for AI/ML-assisted \ac{QoT} estimation \cite{Pointurier:21, Ayassi:22, Allogba:22}.
As a result, \ac{AI/ML}-based \ac{QoT} estimation has become a credible building block for optical network automation.

\replaced{Uncertainty-aware \ac{QoT} methods differ in the predictive uncertainty they represent, the assumptions they require, and the point at which they enter the pipeline. Probabilistic regression~\cite{nicola-22} and \ac{QR}~\cite{somaiieh:23} estimate input-dependent predictive distributions or conditional quantiles under the training distribution. They provide input-dependent intervals but require uncertainty-specific training objectives, and source-domain intervals need not remain calibrated after a technology shift. Bayesian recurrent predictors~\cite{nicola-23} and auxiliary self-confidence outputs~\cite{Tanimura:19} expose model or confidence uncertainty but rely on model-specific inference or architecture changes. Digital-twin sensitivity and explicit uncertainty have also been used for \ac{QoT} optimization and failure prediction~\cite{Zhang:25,Yang:26,Guo:26}, while margin-aware design and robust planning incorporate uncertainty at the network-design level~\cite{Karandin:23,VillavicencioPaz:26}. These approaches suit retrainable predictors or planning workflows but do not provide a model-agnostic finite-sample risk guarantee for a fixed deployed regressor.}{Beyond point prediction, several studies consider uncertainty-aware \ac{AI/ML} methods, including calibrated probabilistic regression~\cite{nicola-22}, Bayesian recurrent forecasting~\cite{nicola-23}, confidence-like outputs~\cite{somaiieh:23}, digital-twin-based \ac{QoT} optimization with sensitivity-guided, memory-aware, and incremental input refinement~\cite{Zhang:25, Yang:26}, trustworthy failure prediction with explicit uncertainty~\cite{Guo:26}, margin- and risk-aware physical-layer design~\cite{Karandin:23}, and robust and adaptive planning~\cite{VillavicencioPaz:26}. However, these studies primarily embed uncertainty within specific models. In contrast, operational decision rules with finite-sample risk control from calibration data remain unexplored in optical network use cases, especially when uncertainty estimates must be translated into decisions without modifying the existing model.}

\replaced{Domain shift mitigation instead targets predictive accuracy through sample-distribution matching~\cite{Gu:23}, evolutionary \acf{TL}~\cite{Zhou:24}, adversarial adaptation~\cite{Cai:24,Housseini_2026_CrossDomainGeneralizationOptical}, or lifelong learning~\cite{Wang:24,Lechowicz_2025_QoTEstimationMarginDriven}. These methods are appropriate when model access and target-domain training are available, but their adapted point estimates alone do not quantify reliability.}{Domain shift mitigation has also attracted attention in \ac{AI/ML}-based optical networking. Existing approaches include sample-distribution matching~\cite{Gu:23}, evolutionary \acf{TL}~\cite{Zhou:24}, adversarial domain adaptation~\cite{Cai:24, Housseini_2026_CrossDomainGeneralizationOptical}, and lifelong learning for time-varying networks~\cite{Wang:24, Lechowicz_2025_QoTEstimationMarginDriven}. While these methods improve target-domain prediction performance, they are often model-specific and rely on retraining or adaptation, which limits their practicality when only limited target-domain data are available after upgrading the network components.}

\replaced{Complementing both families, \ac{CP} attaches distribution-free uncertainty to any fixed predictor using held-out calibration data~\cite{angelopoulos2022gentleintroductionconformalprediction}. Conformalized quantile regression~\cite{Romano-19}, weighted conformal methods under covariate shift~\cite{Tibshirani-19}, and adaptive conformal methods under distribution shift~\cite{Gibbs-21} address heteroscedasticity or changing deployment conditions, while \ac{CRC} extends conformal calibration to user-defined risks~\cite{Angelopoulos:22CRC}. This post-hoc family is suitable when retraining is unavailable but representative labeled calibration data can be collected. The remaining gap is to connect its finite-sample band guarantee to optical network decision support while stating separately which decision errors are guaranteed and which are empirical.}{ Complementing model-specific uncertainty and adaptation methods, \ac{CP} offers a post-hoc way to attach uncertainty estimates to existing predictors, using held-out calibration data to construct prediction sets or intervals that contain the true output with a prescribed probability under standard data assumptions~\cite{angelopoulos2022gentleintroductionconformalprediction}. This model-agnostic property is attractive when different \ac{QoT} regressors are used across studies and deployments. Several extensions target regression and shifted deployment settings, including conformalized quantile regression~\cite{Romano-19}, weighted conformal methods for covariate shift~\cite{Tibshirani-19}, and adaptive conformal methods under distribution shift~\cite{Gibbs-21}, while \ac{CRC} generalizes the conformal idea from covering the true output to controlling broader user-defined risks~\cite{Angelopoulos:22CRC}. However, there remains a gap in using conformal uncertainty for downstream decisions under explicit, finite-sample risk control, especially when source-trained \ac{QoT} models are reused after fiber, amplifier, or switching changes, where only limited calibration data are available in the target domain.}

\begin{figure*}[t]
    \centering
    \includegraphics[width=\textwidth]{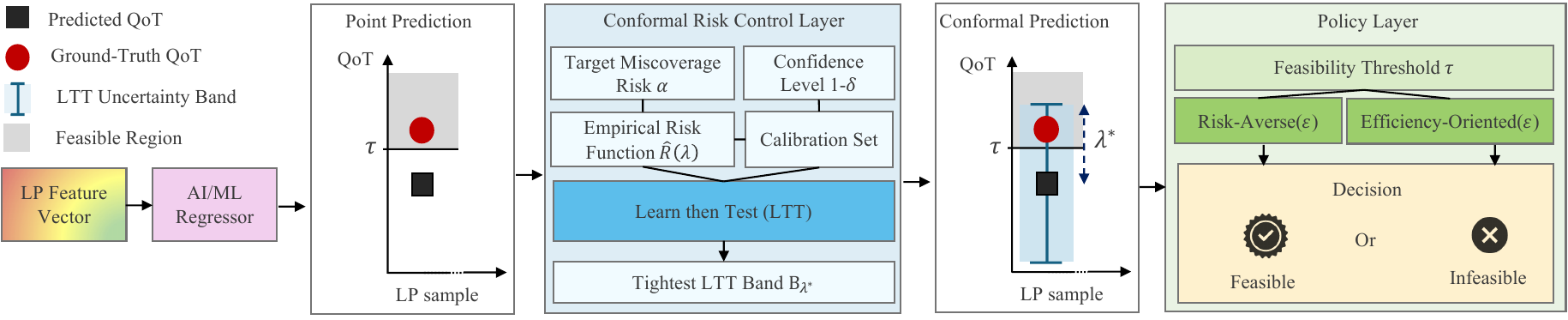}
    \caption{Overview of the proposed \ac{CRC} decision-support framework. An AI/ML regressor predicts the \ac{QoT} of an \ac{LP} from its feature vector. The \ac{CRC} layer uses calibration set and \ac{LTT} to construct the uncertainty band $B_{\lambda^{\star}}$, which is then mapped to a feasibility decision by the \ac{RA} and \ac{EO} policies with respect to the threshold $\tau$.}
    \label{fig:framework}
\end{figure*}

\section{Proposed CRC Decision-Support Framework}
\label{sec:framework}

Figure~\ref{fig:framework} illustrates the proposed \ac{CRC} decision-support framework for \ac{QoT}-aware \ac{LP} feasibility assessment under source--target deployment mismatch.
The framework has three layers:
\emph{i}) a base \ac{AI/ML} regressor that maps an input \ac{LP} feature vector to a point prediction of a \ac{QoT} metric, such as \ac{OSNR} or \ac{BER}, allowing the framework to remain agnostic to the specific regression target;
\emph{ii}) a \ac{CRC} layer that uses held-out target-domain data, referred to as calibration data, and \ac{LTT} to construct a risk-controlled uncertainty band; 
\emph{iii}) a policy layer that maps this band into a feasibility decision according to the desired operational preference, balancing avoidance of infeasible \ac{LP} acceptance against admission of potentially feasible ones.

By separating prediction, band construction, and decision-making, the design lets an existing source-trained \ac{QoT} regressor be reused under deployment mismatch, using limited target-domain data to quantify the uncertainty of its predictions and translating the resulting band into feasibility decisions according to the desired operating mode. The next three subsections describe the framework layers, while the final subsection characterizes the source--target mismatch considered in this work.

%%%%%%%%%% NOTATION GUIDANCE %%%%%%%%%
% X,Y: random feature vector and random true \ac{QoT} value.
% x,y: one realized sample, e.g., a specific \ac{LP}.
% x_i,y_i: calibration samples.
% \hat{f}(x): prediction for one realized input.
% \hat{f}(X): prediction as a random variable for a future random input.
%%%%%%%%%%%%%%%%%%%%%%%%%%%%%%%%%%%%

\subsection{AI/ML Regressor: QoT Estimation}
\label{subsec:base_predictor}

The first layer is an \ac{AI/ML}-based \ac{QoT} regressor that maps an \ac{LP} feature vector $x$ to a point prediction $\hat{y}=\hat{f}(x)$.
We assume the \ac{QoT} metric is monotonic in transmission quality, with larger values better for \ac{OSNR} and smaller values better for \ac{BER}.
The framework treats $\hat{f}$ as a fixed black-box regressor, requiring no change to its architecture, loss, or training procedure, and is therefore applicable to any regression model used for \ac{QoT} estimation.

\subsection{\ac{CRC} Layer: \ac{LTT} Uncertainty Band Construction}
\label{subsec:crc_layer}

The \ac{CRC} layer converts the point prediction of the pretrained \ac{QoT} regressor into a risk-controlled uncertainty band.
The goal is to select a half-band width ($\lambda^{\star}$) that remains as narrow as possible while keeping the probability that the true \ac{QoT} value associated with a future \ac{LP} feature vector falls outside the band below a prescribed level $\alpha$, referred to as the \emph{target miscoverage risk}.

Let $\mathcal{D}_{\mathrm{cal}}=\{(x_i,y_i)\}_{i=1}^{n_{\mathrm{cal}}}$ be a calibration set that is disjoint from the training data, where $x_i$ is an \ac{LP} feature vector and $y_i$ is the corresponding true \ac{QoT} value.
For a candidate half-band width $\lambda\ge 0$, the uncertainty band around the predicted \ac{QoT} value is
\[
B_{\lambda}(x_i)
=
\left[\hat{f}(x_i)-\lambda,\; \hat{f}(x_i)+\lambda\right].
\]
For a calibration \ac{LP} with feature vector $x_i$ and true \ac{QoT} value $y_i$, the band fails to cover the true value when $|y_i-\hat{f}(x_i)|>\lambda$.
Therefore, for each candidate half-band width $\lambda$, we compute
\[
\widehat{R}(\lambda)
=
\frac{1}{n_{\mathrm{cal}}}
\sum_{i=1}^{n_{\mathrm{cal}}}
\mathbf{1}\{|y_i-\hat{f}(x_i)|>\lambda\},
\]
where $\widehat{R}(\lambda)$ is the empirical risk function, i.e., the fraction of calibration \acp{LP} whose true \ac{QoT} values fall outside the candidate uncertainty band.
Selecting $\lambda$ from the condition $\widehat{R}(\lambda)\le\alpha$ would choose a band based only on its observed performance on the available calibration \acp{LP}.

This condition, however, raises two issues.
First, when only limited target-domain data are available after an upgrade, a band that appears sufficient on the calibration \acp{LP} may be too narrow for future test \acp{LP}, leading to overconfident decisions.
Second, since $\lambda^{\star}$ is selected from a candidate grid tested on the same calibration set, the risk-control statement must remain valid after this data-driven selection.
The \ac{LTT} step addresses both by replacing the empirical risk with a concentration upper confidence bound $\widehat{R}^{+}(\lambda)$ on the true miscoverage risk $R(\lambda)$, which holds simultaneously for all candidate half-band widths at confidence level $1-\delta$~\cite{angelopoulos2022learntestcalibratingpredictive}: being uniform over the grid, it absorbs the multiplicity of the selection, and being a high-probability bound rather than an empirical average, it remains meaningful when $n_{\mathrm{cal}}$ is small.
Thus, a candidate half-band width $\lambda$ is accepted only if
\[
\widehat{R}^{+}(\lambda)
\le
\alpha .
\label{eq:accepted_lambda_simple}
\]

\begin{table*}[t]
\centering
\revisedblock
\caption{Evaluated source--target shift families on the CONUS topology
from the SDM \ac{QoT} dataset~\cite{Akbari:25}.
Target-domain sample counts are reported below the table.}
\label{tab:shift_scenarios}
\renewcommand{\arraystretch}{1.08}
\setlength{\tabcolsep}{10pt}

\begin{tabular*}{\textwidth}
{@{\extracolsep{\fill}}lcccl@{}}
\toprule
\textbf{Shift Family}
& \textbf{Fiber}
& \textbf{Amplifier}
& \textbf{Switching}
& \textbf{Source$\rightarrow$Target Scenarios} \\
\midrule

\ac{FAS} shift
& \checkmark & \checkmark & \checkmark
& $3\rightarrow\{7,8\}$ \\

\ac{FA} shift
& \checkmark & \checkmark & --
& $1\rightarrow7;\ 2\rightarrow8;\ 3\rightarrow9$ \\

A shift
& -- & \checkmark & --
& $4\rightarrow7;\ 5\rightarrow8;\ 6\rightarrow9$ \\

F shift
& \checkmark & -- & --
& $1\rightarrow4;\ 2\rightarrow5;\ 3\rightarrow6$ \\

S shift (WC-MCF)
& -- & -- & \checkmark
& $6\rightarrow\{4,5\};\ 9\rightarrow\{7,8\}$ \\

S shift (Bu-SMF)
& -- & -- & \checkmark
& $3\rightarrow\{1,2\}$ \\

\bottomrule
\end{tabular*}

\vspace{3pt}
\begin{minipage}{0.98\textwidth}
\footnotesize
Approximate target-domain calibration/test sample counts
($\times10^3$) are:
$D_1=172/516$,
$D_2=173/518$,
$D_4=166/497$,
$D_5=166/499$,
$D_6=166/497$,
$D_7=176/527$,
$D_8=180/539$, and
$D_9=181/542$.
Here, $D_i$ denotes target dataset $i$ and each pair is reported as
calibration/test.
\end{minipage}
\end{table*}

Among all accepted candidate half-band widths, the \ac{CRC} layer selects the smallest one
\begin{equation}
\lambda^{\star}
=
\min
\left\{
\lambda:
\widehat{R}^{+}(\lambda)
\le
\alpha
\right\}.
\label{eq:lambda_star_simple}
\end{equation}
If no candidate half-band width satisfies this condition, no \ac{LTT} uncertainty band is returned.
Otherwise, for a new test \ac{LP} feature vector $x$, the final \ac{LTT} uncertainty band is
\begin{equation}
B_{\lambda^{\star}}(x)
=
\left[\hat{f}(x)-\lambda^{\star},\; \hat{f}(x)+\lambda^{\star}\right],
\end{equation}
where $\lambda^{\star}$ is the selected \ac{QoT} uncertainty margin around the source-trained regressor prediction.
\replaced{We assume that calibration and future test \acp{LP} are independent samples from the same target-domain distribution, which makes them exchangeable. Under this assumption, with confidence level $1-\delta$, the probability that a future \ac{LP}'s true \ac{QoT} falls outside the selected uncertainty band is no greater than $\alpha$.}{Under the standard exchangeability assumption between calibration and future test \acp{LP}, this means that, with confidence level $1-\delta$, the probability that the true \ac{QoT} value of a future \ac{LP} falls outside the selected uncertainty band is no greater than $\alpha$.}
The \ac{LTT} procedure therefore selects the \replaced{smallest candidate half-band width $\lambda^\star$}{smallest margin} that satisfies the risk-control criterion, avoiding unnecessary conservatism while providing explicit miscoverage-risk control \added{for the uncertainty band used by} downstream decision tasks.

\subsection{Policy layer: Feasibility Decision}
\label{subsec:policy_layer}

The policy layer defines how the \ac{LTT} band is used to accept or reject an \ac{LP}, supporting different operating modes by applying the feasibility test to either its lower or upper bound.
\deleted{The same \ac{LTT} uncertainty band can support different operating modes by applying the feasibility test to either its lower or upper bound, as formalized next.}
Let
\(
    L(x)=\hat{f}(x)-\lambda^{\star},
    \qquad
    U(x)=\hat{f}(x)+\lambda^{\star}
\)
denote the lower and upper bounds of the band for an \ac{LP} feature vector $x$, and let $\tau$ denote the modulation-format-specific \ac{QoT} threshold.

We consider two policies.
The \ac{RA} policy declares an \ac{LP} feasible only if
\begin{equation}
\label{eq:ra}
    L(x) \ge \tau + \epsilon_{\mathrm{RA}} .
\end{equation}
This policy is conservative, as the prediction must remain above the threshold even at the lower end of the band, and is suitable when avoiding the acceptance of an infeasible \ac{LP} is the main objective.

The \ac{EO} policy declares an \ac{LP} feasible only if
\begin{equation}
\label{eq:eo}
    U(x) \ge \tau + \epsilon_{\mathrm{EO}} .
\end{equation}
This policy is less conservative, accepting an \ac{LP} when the upper end of the band indicates possible feasibility, and is suitable when the operator prioritizes higher-spectral-efficiency modulation formats.
The parameters $\epsilon_{\mathrm{RA}}$ and $\epsilon_{\mathrm{EO}}$ adjust the decision boundary around $\tau$.

\added{The \ac{CRC} guarantee controls the target miscoverage risk at level $\alpha$. Under \ac{RA}, band coverage and the non-negative $\epsilon_{\mathrm{RA}}$ ensure that an \ac{LP} declared feasible is truly feasible. Declaring an infeasible \ac{LP} as feasible requires its actual \ac{QoT} to fall below the band, and the probability of this outcome is no greater than $\alpha$. Under \ac{EO}, declaring an \ac{LP} infeasible despite its actual \ac{QoT} reaching the threshold, including the $\epsilon_{\mathrm{EO}}$, requires its actual \ac{QoT} to exceed the band and is likewise bounded by $\alpha$. These bounds concern different outcomes. Hence, the policies do not provide the same operational guarantee as the uncertainty bands. Their overall performance is evaluated empirically in Section~\ref{subsec:feasibility_decision}.}

\subsection{Source--Target Shift Characterization}
\label{subsec:source_target_shift}

Technology changes may alter the relationship between the \ac{LP} feature vector and the true \ac{QoT} value.
Let $\mathbb{P}_S$ and $\mathbb{P}_T$ denote the source- and target-domain joint distributions of $(X,Y)$, where $X\in\mathbb{R}^{d}$ is the \ac{LP} feature vector and $Y\in\mathbb{R}$ the corresponding \ac{QoT} value.
The mismatch can appear in two ways: the feature values themselves may change (e.g., different path lengths, link occupations, carrier frequencies, or active-channel patterns), and their relation to \ac{QoT} may change after a fiber, amplifier, or switching change.
The shift score below reflects both effects.

\begin{figure*}[t]
    \centering
    \begin{tikzpicture}[x=0.050\textwidth, y=-0.015\textheight]
\definecolor{heat0}{rgb}{0.1294,0.4000,0.6745}
\definecolor{heat1}{rgb}{0.1916,0.4421,0.6963}
\definecolor{heat2}{rgb}{0.2537,0.4842,0.7181}
\definecolor{heat3}{rgb}{0.3159,0.5264,0.7399}
\definecolor{heat4}{rgb}{0.3781,0.5685,0.7617}
\definecolor{heat5}{rgb}{0.4402,0.6106,0.7834}
\definecolor{heat6}{rgb}{0.5024,0.6527,0.8052}
\definecolor{heat7}{rgb}{0.5646,0.6948,0.8270}
\definecolor{heat8}{rgb}{0.6267,0.7370,0.8488}
\definecolor{heat9}{rgb}{0.6889,0.7791,0.8706}
\definecolor{heat10}{rgb}{0.7511,0.8212,0.8924}
\definecolor{heat11}{rgb}{0.8132,0.8633,0.9142}
\definecolor{heat12}{rgb}{0.8754,0.9054,0.9359}
\definecolor{heat13}{rgb}{0.9375,0.9476,0.9577}
\definecolor{heat14}{rgb}{0.9586,0.9362,0.9390}
\definecolor{heat15}{rgb}{0.9386,0.8715,0.8797}
\definecolor{heat16}{rgb}{0.9185,0.8067,0.8205}
\definecolor{heat17}{rgb}{0.8985,0.7419,0.7612}
\definecolor{heat18}{rgb}{0.8784,0.6771,0.7020}
\definecolor{heat19}{rgb}{0.8584,0.6123,0.6427}
\definecolor{heat20}{rgb}{0.8383,0.5476,0.5834}
\definecolor{heat21}{rgb}{0.8183,0.4828,0.5242}
\definecolor{heat22}{rgb}{0.7983,0.4180,0.4649}
\definecolor{heat23}{rgb}{0.7782,0.3532,0.4057}
\definecolor{heat24}{rgb}{0.7582,0.2885,0.3464}
\definecolor{heat25}{rgb}{0.7381,0.2237,0.2871}
\definecolor{heat26}{rgb}{0.7181,0.1589,0.2279}
\definecolor{heat27}{rgb}{0.6980,0.0941,0.1686}

\def\drawheatrow#1#2{%
  \foreach \v [count=\c from 0] in {#2}{%
    \fill[heat\v] (\c,#1) rectangle ++(1,1);
    \draw[white,line width=0.12pt] (\c,#1) rectangle ++(1,1);
  }%
}

% Averaged rows by shift family.
\drawheatrow{0}{5,1,2,2,16,20,22,14,4,5,1,2,23,22} % FAS-Shift
\drawheatrow{1}{5,1,2,2,9,13,14,11,5,5,2,2,9,14}  % FA-Shift
\drawheatrow{2}{7,2,2,3,10,16,16,13,7,7,2,2,11,18} % A-Shift
\drawheatrow{3}{0,0,0,0,0,0,0,0,0,0,0,0,0,0}       % F-Shift
\drawheatrow{4}{0,0,0,0,1,1,1,0,0,0,0,0,2,1}       % S-Shift (WC-MCF)
\drawheatrow{5}{0,0,0,0,1,1,1,0,0,0,0,0,2,1}       % S-Shift (Bu-SMF)

% More visible feature-category separators.
\draw[white,line width=2.2pt] (4,0) -- (4,6);
\draw[white,line width=2.2pt] (8,0) -- (8,6);
\draw[white,line width=2.2pt] (12,0) -- (12,6);

% Outer frame.
\draw[black!20,line width=0.35pt] (0,0) rectangle (14,6);

%\foreach \x/\lab in {
%0.5/Path,1.5/Avg.,2.5/Min.,3.5/Max.,
%4.5/Min.,5.5/Max.,6.5/Avg.,7.5/Std.,
%8.5/{\# of Links},9.5/{\# of Spans},
%10.5/{Source Node Deg.},11.5/{Dest. Node Deg.},
%12.5/{Carrier Freq.},13.5/{Active Channels}
%}{
%  \node[rotate=45,anchor=east,font=\footnotesize] at (\x,6.20) {\lab};
%}
\foreach \x/\lab in {
0.5/PL,
1.5/LL-Avg.,2.5/LL-Min.,3.5/LL-Max.,
4.5/LO-Min.,5.5/LO-Max.,6.5/LO-Avg.,7.5/LO-Std.,
8.5/NL,9.5/NS,
10.5/SD,11.5/DD,
12.5/CF,13.5/AC
}{
  \node[rotate=45,anchor=east,font=\footnotesize] at (\x,6.20) {\lab};
}

\foreach \y/\lab in {
0.5/{FAS shift},
1.5/{FA shift},
2.5/{A shift},
3.5/{F shift},
4.5/{S shift (WC-MCF)},
5.5/{S shift (Bu-SMF)}
}{
  \node[anchor=east,font=\footnotesize,align=right] at (-0.12,\y) {\lab};
}

\node[anchor=south,font=\footnotesize] at (2,-0.10) {Length};
\node[anchor=south,font=\footnotesize] at (6,-0.10) {Link occupation};
\node[anchor=south,font=\footnotesize] at (10,-0.10) {Topology};
\node[anchor=south,font=\footnotesize] at (13,-0.10) {Spectrum};

% Vertical colorbar on the right: low at bottom, high at top.
\foreach \i in {0,...,27}{
  \pgfmathsetmacro{\y}{5.8 - 0.20*\i}
  \fill[heat\i] (15.05,\y) rectangle ++(0.30,-0.20);
}
\draw[black!30] (15.05,0.20) rectangle (15.35,5.80);

%\node[anchor=west,font=\scriptsize] at (15.48,5.80) {0.000};
%\node[anchor=west,font=\scriptsize] at (15.48,0.20) {0.175};

%\node[anchor=south,font=\footnotesize,rotate=90] at (15.90,3.00) %{Shift Contribution Score};
%{SCS};

\node[anchor=south,font=\footnotesize,align=center] at (15.20,-0.15) %{Shift\\Contribution Score};
{$C_j$};
\node[anchor=west,font=\scriptsize] at (15.48,5.80) {0.000};
\node[anchor=west,font=\scriptsize] at (15.48,0.20) {0.175};
\end{tikzpicture}
    %\caption{Shift heatmap.}
   \caption{Average feature shift heatmap of $C_j$ across source--target shift families. Features are grouped by the characteristic they represent: length, link occupation, topology, and spectrum. Feature abbreviations: PL = path length, LL = link length, LO = link occupation, NL = number of links, NS = number of spans, SD = source node degree, DD = destination node degree, CF = carrier frequency, and AC = active channels.}
    \label{fig:shift_feature_heatmap}
\end{figure*}
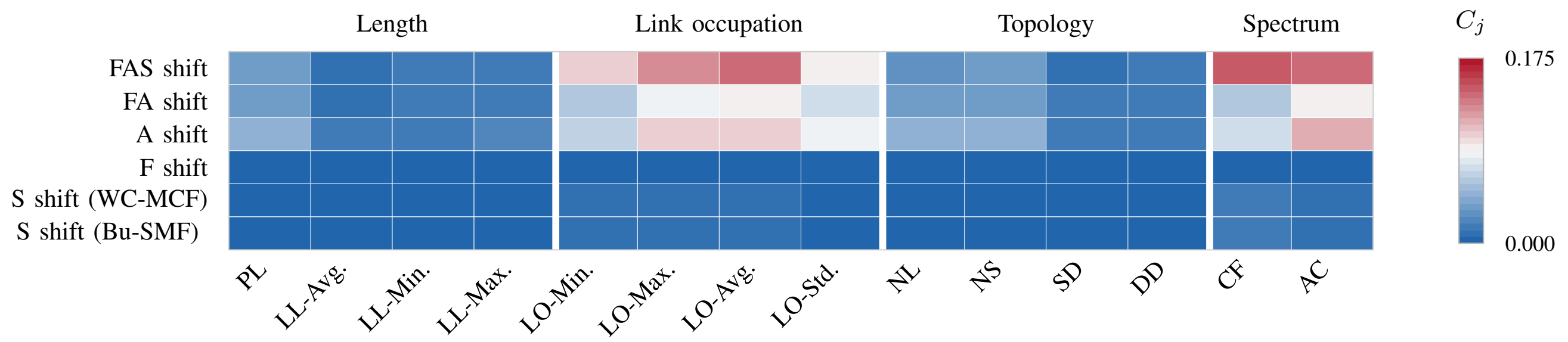
To measure how much an individual feature changes between domains, we use the 1D Wasserstein distance ($W_1$)~\cite{Villani2009OT, Cai:24}, a standard metric for comparing empirical distributions, applied one feature at a time.
Letting $X^{(j)}$ denote the $j$th feature, we define a feature-shift indicator as
\[
D_j^{X}
=
W_1\!\left(\mathbb{P}_S(X^{(j)}),\mathbb{P}_T(X^{(j)})\right).
\]
Similarly, to capture changes in the relation between the $j$th feature and the \ac{QoT} value across domains, we define a conditional-shift indicator $D_j^{Y|X}$, which measures how the \ac{QoT} distribution differs between source and target domains for comparable values of the $j$th feature.
We divide the range of $X^{(j)}$ into $K$ bins $\{\mathcal{B}^{j}_{k}\}_{k=1}^{K}$ and define
\[
D_j^{Y|X}
=
\frac{1}{K}
\sum_{k=1}^{K}
W_1\!\left(
\mathbb{P}_S(Y {\mid} X^{(j)} \in \mathcal{B}^{j}_{k}),
\mathbb{P}_T(Y {\mid} X^{(j)} \in \mathcal{B}^{j}_{k})
\right),
\]
where $\mathcal{B}^{j}_{k}$ is the $k$th bin of the $j$th feature.
Thus, $D_j^{Y|X}$ is large when \acp{LP} with comparable values of the $j$th feature exhibit different \ac{QoT} behavior across domains.
The $j$th feature shift contribution score is then
\begin{equation}
C_j
=
D_j^{X}D_j^{Y|X}.
\label{eq:shift_contribution}
\end{equation}
A large $C_j$ indicates that the $j$th feature changes across domains and that its relation to \ac{QoT} also changes.
We use $C_j$ to explain which \ac{LP} features contribute most to the mismatch.
It is not used for model adaptation and is not a formal distance between the full distributions. Being a product, it also does not detect a transition that alters the feature--\ac{QoT} relation while leaving the feature distributions intact, so it must be analyzed alongside the selected band width.

\section{Experimental Setup}
\label{sec:exp_setup}

We use datasets 1--9 from the open \ac{SDM} \ac{QoT} dataset~\cite{Akbari:25} on the CONUS topology, which provide nine technology-specific configurations combining \ac{Bu-SMF} and \ac{WC-MCF} fiber, \ac{EDFA}- and \ac{SDM}-amplified systems, and \ac{J-Sw}, \ac{FrJ-Sw}, and \ac{Ind-Sw} switching.
Taking \ac{OSNR} as the target \ac{QoT} metric, we define the 17 source--target scenarios of Table~\ref{tab:shift_scenarios}, grouped into six domain shift families that cover both single-dimension and compounded technology changes.

Table~\ref{tab:shift_scenarios} summarizes the evaluated source--target scenarios, where F, A, and S denote fiber, amplifier, and switching technology changes, respectively. 
The scenarios include simultaneous three-dimensional changes (\ac{FAS} shift), joint fiber--amplifier changes (\ac{FA} shift), and single-dimension changes (F, A, and S shift), reflecting practical network upgrades under a fixed network footprint.
For each scenario, the source--target mismatch is characterized using the feature shift contribution score $C_j$ defined in~\eqref{eq:shift_contribution}, which is used to interpret the severity and origin of the domain shift.

\added{For each scenario, source-domain data train an \ac{XGB} \ac{QoT} regressor. The 20\% target adaptation subset is fixed, while the remaining target data are independently/randomly repartitioned ten times into disjoint 20\% calibration and 60\% test subsets. For each partition, \ac{CRC} uses $\alpha=0.10$, $1-\delta=0.95$, and 60 uniformly spaced candidate half-band widths $\lambda$ between $0$ and $1.3Q_{0.95}$, where $Q_{0.95}$ is the 95th percentile of the absolute calibration residuals. The selected band is then evaluated on the corresponding test split, yielding 170 trials. Overlapping test samples assess partition sensitivity rather than independent-dataset variation; Table~\ref{tab:shift_scenarios} gives rounded sample counts, while exact scenario-specific counts and experiment seeds are provided in Supplementary Table~S1 and Supplementary Section~I.}
\deleted{For each scenario, the source-domain data train an \ac{XGB}-based \ac{QoT} regressor that serves as the base predictor, and the target-domain data are split into three disjoint subsets: 20\% for calibration, 20\% reserved for adaptation, and 60\% for testing, kept fixed across scenarios for a fair comparison.
The calibration set is used only by the \ac{CRC} layer to select the smallest half-band width $\lambda^{\star}$ under the target miscoverage risk $\alpha=0.1$ and confidence level $1-\delta=0.95$, while the adaptation subset is used only by the \ac{TL} policy.}

\added{We compare source-only \ac{ZS}, target-adapted \ac{TL} trained on the adaptation subset, and the proposed \ac{RA} and \ac{EO} policies. As a separate uncertainty-quantification baseline, we adapt the \ac{QR} approach of~\cite{somaiieh:23} to our source-domain setting. Each endpoint, $\widehat q_{0.05}(x)$ or $\widehat q_{0.95}(x)$, is the mean of five independently trained \ac{XGB} quantile models to reduce seed-induced variance. The interval uses neither target training nor calibration and is evaluated on the same test samples as \ac{CRC}; coverage and width are compared under technology shifts.}
\deleted{We compare the proposed \ac{RA} and \ac{EO} policies against two point-prediction baselines that decide solely from point \ac{OSNR} predictions without the \ac{CRC} layer: the \ac{ZS} policy applies the source-trained regressor directly to the target data, while the \ac{TL} policy fine-tunes it on the reserved adaptation subset and provides a practical upper reference.}

\added{Calibration-size sensitivity fixes each test split and recalibrates on nested subsets of 500, 1,000, 2,500, 5,000, 10,000, 25,000, 50,000, and 100,000 samples, plus the full calibration set, across 1,530 scenario--partition runs.}
\added{For policy evaluation, $\tau$ is set to 12, 14, 17, 21, and 24~dB for QPSK, 8-QAM, 16-QAM, 32-QAM, and 64-QAM, respectively. Modulation-specific \ac{OSNR} thresholding follows common elastic optical-network planning practice~\cite{Mitra:16}. We use $\epsilon_{\mathrm{RA}}=0$~dB and $\epsilon_{\mathrm{EO}}=0.1$~dB: \ac{RA} relies on the lower bound for conservatism, whereas \ac{EO} requires 0.1~dB headroom above $\tau$. These fixed settings are applied to all scenarios.}

Since \ac{LP} feasibility assessment is a binary classification task, performance is measured using accuracy, F1-score, \acf{FPR}, and \acf{FNR}, with \ac{FPR} corresponding to infeasible \acp{LP} incorrectly labeled as feasible and \ac{FNR} corresponding to feasible \acp{LP} incorrectly rejected as infeasible.
These two error types capture distinct operational consequences, since \ac{FPR} may lead to failed \ac{LP} establishment or service disruption, whereas \ac{FNR} may reduce provisioning opportunities or resource efficiency.

\begin{table*}[t]
\centering
\revisedblock
\caption{Test coverage and full interval width for \ac{LTT} uncertainty band and source-only \ac{QR}, by shift family. Each scenario is first averaged over ten target-data partitions; values are mean $\pm$ standard deviation across scenarios.}
\label{tab:ltt-coverage-family-zs}
\renewcommand{\arraystretch}{1.08}
\setlength{\tabcolsep}{10pt}
\small
\begin{tabular}{lcccc}
\toprule
& \multicolumn{2}{c}{\textbf{\ac{LTT} uncertainty band}}
& \multicolumn{2}{c}{\textbf{Source-only \ac{QR} band}} \\
\cmidrule(lr){2-3}\cmidrule(lr){4-5}
\textbf{Shift Family}
& \textbf{Test cov. (\%)}
& \textbf{Full width [dB]}
& \textbf{Test cov. (\%)}
& \textbf{Full width [dB]} \\
\midrule
\ac{FAS} shift & $91.65 \pm 0.11$ & $4.200 \pm 0.027$ & $0.00 \pm 0.00$ & $0.219 \pm 0.000$ \\
\ac{FA} shift & $91.49 \pm 0.37$ & $4.161 \pm 0.017$ & $0.00 \pm 0.00$ & $0.221 \pm 0.003$ \\
A shift & $94.43 \pm 3.72$ & $4.663 \pm 0.041$ & $0.00 \pm 0.00$ & $0.216 \pm 0.004$ \\
F shift & $91.14 \pm 0.28$ & $0.699 \pm 0.026$ & $3.28 \pm 0.45$ & $0.215 \pm 0.003$ \\
S shift (WC-MCF) & $91.03 \pm 0.09$ & $0.178 \pm 0.053$ & $82.70 \pm 3.66$ & $0.182 \pm 0.025$ \\
S shift (Bu-SMF) & $91.05 \pm 0.12$ & $0.235 \pm 0.021$ & $82.16 \pm 4.71$ & $0.215 \pm 0.000$ \\
\bottomrule
\end{tabular}

\vspace{2pt}
{\footnotesize %\ac{CRC}: $\alpha=0.10$, $1-\delta=0.95$. 
\ac{QR}: nominal 90\% source-domain interval with no target-domain calibration.}
\end{table*}

\begin{figure*}[t]
    \centering
    %\vspace{-2mm}

    \ref{sharedbandslegend}
    %\vspace{-1.5mm}

    \begin{subfigure}[t]{0.32\textwidth}
        \centering
        % Auto-generated by qot_sdm/revision_experiments/generate_crc_qr_band_figure3.py.
% The original 15 CRC figure lightpaths are preserved; QR is evaluated on the same rows.
\begin{tikzpicture}
\begin{axis}[
    width=\linewidth,
    height=0.50\linewidth,
    ymin=17.5236,
    ymax=24.5323,
    xmin=-0.5,
    xmax=14.5,
    xlabel={Selected test \acp{LP}},
    ylabel={OSNR [dB]},
    axis x line*=bottom,
    axis y line*=left,
    xtick pos=bottom,
    ytick pos=left,
    tick align=outside,
    x tick label style={font=\footnotesize},
    y tick label style={font=\footnotesize},
    label style={font=\footnotesize},
    xlabel style={font=\footnotesize, yshift=1pt},
    ylabel style={font=\footnotesize, yshift=-2pt},
    ymajorgrids,
    grid style={black!12},
    clip=false,
    legend to name=sharedbandslegend,
    legend columns=4,
    legend style={
        draw=none,
        font=\scriptsize,
        text=black,
        /tikz/every even column/.append style={column sep=0.22cm}
    },
]

% Original LTT-calibrated CRC band and source-model prediction.
\addplot[draw=none, mark=none, fill=LttBlue, fill opacity=0.24, forget plot]
coordinates {(0,22.079834) (1,22.155989) (2,22.267258) (3,22.337904) (4,22.638674) (5,22.482697) (6,22.517334) (7,22.617184) (8,22.592072) (9,22.778670) (10,22.829758) (11,23.163506) (12,23.625565) (13,23.695824) (14,24.156815) (14,19.976051) (13,19.515060) (12,19.444801) (11,18.982742) (10,18.648994) (9,18.597906) (8,18.411308) (7,18.436420) (6,18.336570) (5,18.301933) (4,18.457910) (3,18.157140) (2,18.086494) (1,17.975225) (0,17.899070)} \closedcycle;

% Source-only QR band evaluated on the identical original lightpaths.
\addplot[draw=none, mark=none,
        fill=OrangeBand, fill opacity=0.38, forget plot]
coordinates {(0,20.140289) (1,20.214340) (2,20.315748) (3,20.351206) (4,20.639847) (5,20.532503) (6,20.553181) (7,20.645056) (8,20.625641) (9,20.795938) (10,20.827618) (11,21.177635) (12,21.653456) (13,21.682686) (14,22.142330) (14,21.856590) (13,21.455227) (12,21.462715) (11,20.961548) (10,20.642117) (9,20.585844) (8,20.380295) (7,20.421791) (6,20.314253) (5,20.325867) (4,20.399761) (3,20.132957) (2,20.102325) (1,19.966267) (0,19.888050)} \closedcycle;
\addplot[OrangeBand!85!black, densely dotted, line width=0.55pt, mark=none,
         forget plot] coordinates {(0,19.888050) (1,19.966267) (2,20.102325) (3,20.132957) (4,20.399761) (5,20.325867) (6,20.314253) (7,20.421791) (8,20.380295) (9,20.585844) (10,20.642117) (11,20.961548) (12,21.462715) (13,21.455227) (14,21.856590)};
\addplot[OrangeBand!85!black, densely dotted, line width=0.55pt, mark=none,
         forget plot] coordinates {(0,20.140289) (1,20.214340) (2,20.315748) (3,20.351206) (4,20.639847) (5,20.532503) (6,20.553181) (7,20.645056) (8,20.625641) (9,20.795938) (10,20.827618) (11,21.177635) (12,21.653456) (13,21.682686) (14,22.142330)};

\addlegendimage{area legend, draw=LttBlue, fill=LttBlue, fill opacity=0.24}
\addlegendentry{\textcolor{black}{\ac{LTT} uncertainty band}}
\addlegendimage{area legend, draw=OrangeBand!85!black, densely dotted,
                 fill=OrangeBand, fill opacity=0.38}
\addlegendentry{\textcolor{black}{\ac{QR} band}}

% True target-domain OSNR.
\addplot[mark=*, mark size=1.05pt, color=TruthRed,
         mark options={draw=TruthRed, fill=TruthRed}, line width=0.55pt]
coordinates {(0,22.049521) (1,22.168117) (2,22.182486) (3,22.305418) (4,22.458919) (5,22.469209) (6,22.488759) (7,22.526949) (8,22.549036) (9,22.688917) (10,22.780739) (11,23.088184) (12,23.483277) (13,23.586129) (14,23.665710)};
\addlegendentry{\textcolor{black}{True \ac{OSNR}}}

% Fixed source-regressor prediction used by CRC.
\addplot[mark=square*, mark size=0.95pt, color=PredBlack,
         mark options={draw=PredBlack, fill=PredBlack}, dashed, line width=0.55pt]
coordinates {(0,19.989452) (1,20.065607) (2,20.176876) (3,20.247522) (4,20.548292) (5,20.392315) (6,20.426952) (7,20.526802) (8,20.501690) (9,20.688288) (10,20.739376) (11,21.073124) (12,21.535183) (13,21.605442) (14,22.066433)};
\addlegendentry{\textcolor{black}{Source prediction}}
\end{axis}
\end{tikzpicture}
        \caption{\ac{FAS} shift}
        \label{fig:bands-fas-3-8}
    \end{subfigure}
    \hfill
    \begin{subfigure}[t]{0.32\textwidth}
        \centering
        % Auto-generated by qot_sdm/revision_experiments/generate_crc_qr_band_figure3.py.
% The original 15 CRC figure lightpaths are preserved; QR is evaluated on the same rows.
\begin{tikzpicture}
\begin{axis}[
    width=\linewidth,
    height=0.50\linewidth,
    ymin=17.2843,
    ymax=24.2048,
    xmin=-0.5,
    xmax=14.5,
    xlabel={Selected test \acp{LP}},
    ylabel={OSNR [dB]},
    axis x line*=bottom,
    axis y line*=left,
    xtick pos=bottom,
    ytick pos=left,
    tick align=outside,
    x tick label style={font=\footnotesize},
    y tick label style={font=\footnotesize},
    label style={font=\footnotesize},
    xlabel style={font=\footnotesize, yshift=1pt},
    ylabel style={font=\footnotesize, yshift=-2pt},
    ymajorgrids,
    grid style={black!12},
    clip=false,
]

% Original LTT-calibrated CRC band and source-model prediction.
\addplot[draw=none, mark=none, fill=LttBlue, fill opacity=0.24, forget plot]
coordinates {(0,22.364426) (1,22.495640) (2,22.499739) (3,22.796970) (4,22.836912) (5,23.150988) (6,22.917128) (7,23.016470) (8,23.014500) (9,23.252058) (10,23.478636) (11,23.534770) (12,23.834078) (13,23.793310) (14,23.729977) (14,19.020607) (13,19.083940) (12,19.124708) (11,18.825400) (10,18.769266) (9,18.542688) (8,18.305130) (7,18.307100) (6,18.207758) (5,18.441618) (4,18.127542) (3,18.087600) (2,17.790369) (1,17.786270) (0,17.655056)} \closedcycle;

% Source-only QR band evaluated on the identical original lightpaths.
\addplot[draw=none, mark=none,
        fill=OrangeBand, fill opacity=0.38, forget plot]
coordinates {(0,20.126453) (1,20.238874) (2,20.299892) (3,20.489370) (4,20.633661) (5,20.827745) (6,20.666372) (7,20.788576) (8,20.745684) (9,20.975325) (10,21.253540) (11,21.392227) (12,21.575602) (13,21.556404) (14,21.521149) (14,21.252287) (13,21.358669) (12,21.286398) (11,21.053984) (10,20.964668) (9,20.795811) (8,20.551170) (7,20.559183) (6,20.446564) (5,20.709223) (4,20.409939) (3,20.328379) (2,20.011168) (1,20.044689) (0,19.915304)} \closedcycle;
\addplot[OrangeBand!85!black, densely dotted, line width=0.55pt, mark=none,
         forget plot] coordinates {(0,19.915304) (1,20.044689) (2,20.011168) (3,20.328379) (4,20.409939) (5,20.709223) (6,20.446564) (7,20.559183) (8,20.551170) (9,20.795811) (10,20.964668) (11,21.053984) (12,21.286398) (13,21.358669) (14,21.252287)};
\addplot[OrangeBand!85!black, densely dotted, line width=0.55pt, mark=none,
         forget plot] coordinates {(0,20.126453) (1,20.238874) (2,20.299892) (3,20.489370) (4,20.633661) (5,20.827745) (6,20.666372) (7,20.788576) (8,20.745684) (9,20.975325) (10,21.253540) (11,21.392227) (12,21.575602) (13,21.556404) (14,21.521149)};

% True target-domain OSNR.
\addplot[mark=*, mark size=1.05pt, color=TruthRed,
         mark options={draw=TruthRed, fill=TruthRed}, line width=0.55pt]
coordinates {(0,22.233617) (1,22.307976) (2,22.444010) (3,22.634150) (4,22.722470) (5,22.829607) (6,22.842550) (7,22.851317) (8,22.985157) (9,23.140647) (10,23.370811) (11,23.463721) (12,23.604972) (13,23.619028) (14,23.650495)};

% Fixed source-regressor prediction used by CRC.
\addplot[mark=square*, mark size=0.95pt, color=PredBlack,
         mark options={draw=PredBlack, fill=PredBlack}, dashed, line width=0.55pt]
coordinates {(0,20.009740) (1,20.140955) (2,20.145054) (3,20.442286) (4,20.482227) (5,20.796303) (6,20.562443) (7,20.661785) (8,20.659815) (9,20.897373) (10,21.123951) (11,21.180084) (12,21.479393) (13,21.438625) (14,21.375292)};

\end{axis}
\end{tikzpicture}
        \caption{A shift}
        \label{fig:bands-a-6-9}
    \end{subfigure}
    \hfill
    \begin{subfigure}[t]{0.32\textwidth}
        \centering
        % Auto-generated by qot_sdm/revision_experiments/generate_crc_qr_band_figure3.py.
% The original 15 CRC figure lightpaths are preserved; QR is evaluated on the same rows.
\begin{tikzpicture}
\begin{axis}[
    width=\linewidth,
    height=0.50\linewidth,
    ymin=21.7942,
    ymax=24.2715,
    xmin=-0.5,
    xmax=14.5,
    xlabel={Selected test \acp{LP}},
    ylabel={OSNR [dB]},
    axis x line*=bottom,
    axis y line*=left,
    xtick pos=bottom,
    ytick pos=left,
    tick align=outside,
    x tick label style={font=\footnotesize},
    y tick label style={font=\footnotesize},
    label style={font=\footnotesize},
    xlabel style={font=\footnotesize, yshift=1pt},
    ylabel style={font=\footnotesize, yshift=-2pt},
    ymajorgrids,
    grid style={black!12},
    clip=false,
]

% Original LTT-calibrated CRC band and source-model prediction.
\addplot[draw=none, mark=none, fill=LttBlue, fill opacity=0.24, forget plot]
coordinates {(0,22.775670) (1,22.765830) (2,23.066092) (3,22.905940) (4,23.161870) (5,23.116282) (6,23.146132) (7,23.221972) (8,23.358212) (9,23.275300) (10,23.472270) (11,23.396502) (12,23.632390) (13,23.775670) (14,24.021502) (14,23.299840) (13,23.054008) (12,22.910728) (11,22.674840) (10,22.750608) (9,22.553638) (8,22.636550) (7,22.500310) (6,22.424470) (5,22.394620) (4,22.440208) (3,22.184278) (2,22.344430) (1,22.044168) (0,22.054008)} \closedcycle;

% Source-only QR band evaluated on the identical original lightpaths.
\addplot[draw=none, mark=none,
        fill=OrangeBand, fill opacity=0.38, forget plot]
coordinates {(0,22.524349) (1,22.512569) (2,22.818794) (3,22.697706) (4,22.895327) (5,22.827778) (6,22.866270) (7,23.085735) (8,23.150507) (9,22.945221) (10,23.241678) (11,23.141075) (12,23.361408) (13,23.696291) (14,23.799757) (14,23.517326) (13,23.265495) (12,23.200359) (11,22.873489) (10,22.954987) (9,22.791754) (8,22.881681) (7,22.710056) (6,22.681274) (5,22.603523) (4,22.627495) (3,22.439465) (2,22.571110) (1,22.311630) (0,22.329094)} \closedcycle;
\addplot[OrangeBand!85!black, densely dotted, line width=0.55pt, mark=none,
         forget plot] coordinates {(0,22.329094) (1,22.311630) (2,22.571110) (3,22.439465) (4,22.627495) (5,22.603523) (6,22.681274) (7,22.710056) (8,22.881681) (9,22.791754) (10,22.954987) (11,22.873489) (12,23.200359) (13,23.265495) (14,23.517326)};
\addplot[OrangeBand!85!black, densely dotted, line width=0.55pt, mark=none,
         forget plot] coordinates {(0,22.524349) (1,22.512569) (2,22.818794) (3,22.697706) (4,22.895327) (5,22.827778) (6,22.866270) (7,23.085735) (8,23.150507) (9,22.945221) (10,23.241678) (11,23.141075) (12,23.361408) (13,23.696291) (14,23.799757)};

% True target-domain OSNR.
\addplot[mark=*, mark size=1.05pt, color=TruthRed,
         mark options={draw=TruthRed, fill=TruthRed}, line width=0.55pt]
coordinates {(0,22.088351) (1,22.137461) (2,22.367000) (3,22.389666) (4,22.444334) (5,22.501499) (6,22.512536) (7,22.567842) (8,22.706995) (9,22.710668) (10,22.802932) (11,22.874514) (12,23.173001) (13,23.223288) (14,23.387282)};

% Fixed source-regressor prediction used by CRC.
\addplot[mark=square*, mark size=0.95pt, color=PredBlack,
         mark options={draw=PredBlack, fill=PredBlack}, dashed, line width=0.55pt]
coordinates {(0,22.414839) (1,22.404999) (2,22.705261) (3,22.545109) (4,22.801039) (5,22.755451) (6,22.785301) (7,22.861141) (8,22.997381) (9,22.914469) (10,23.111439) (11,23.035671) (12,23.271559) (13,23.414839) (14,23.660671)};

\end{axis}
\end{tikzpicture}
        \caption{F shift}
        \label{fig:bands-f-3-6}
    \end{subfigure}
    \caption{\replaced{\ac{LTT} uncertainty and source-only \ac{QR} bands for the same 15 target-domain test \acp{LP} and selected shifts.}{\ac{LTT} uncertainty bands around predicted \ac{OSNR} values, shown together with the true \ac{OSNR} values, for 15 representative test \acp{LP} under selected domain shift families in the \ac{ZS} setting.}}
    \label{fig:bands-selected-shifts-zs}
\end{figure*}
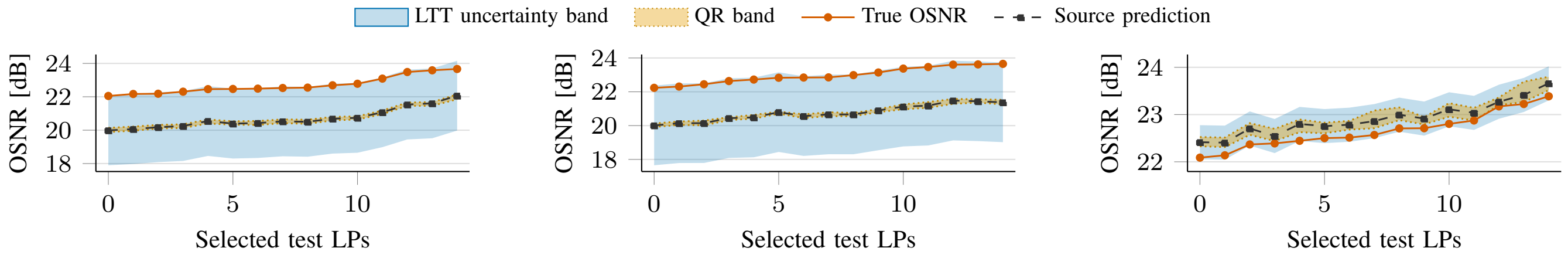

\section{Results and Discussion}
\label{sec:results}

We first characterize dataset shift using feature shift contribution score $C_j$ defined in Section \ref{subsec:source_target_shift}, then \replaced{compare the target-domain coverage and width of the \ac{LTT} uncertainty and source-only \ac{QR} bands.}{verify \ac{LTT} empirical coverage to assess whether the selected bands generalize to unseen target-domain \acp{LP}.}
For \ac{LP} feasibility assessment, we evaluate the proposed \ac{RA} and \ac{EO} policies against \ac{ZS} and \ac{TL} point-prediction baseline policies on the same held-out test \replaced{samples}{subset}.

\begin{figure*}[t]
    \centering

    \begin{subfigure}{\textwidth}
        \centering
        % Figure 4(a): test-set means over ten partitions per scenario,
% then averaged across scenarios within each shift family.
% ZS and TL models remain fixed; RA and EO use the full calibration sets.
% Requires qot_sdm_plot_preamble.tex in the manuscript preamble.
\begin{tikzpicture}
\begin{axis}[
    width=0.98\textwidth,
    height=0.22\textheight,
    ybar,
    ymin=60,
    ymax=103,
    ylabel={Rate (\%)},
    ytick pos=left,
    yticklabel style={font=\footnotesize},
    xtick={0.0000,1.0000,2.5500,3.5500,5.1000,6.1000,7.6500,8.6500,10.2000,11.2000,12.7500,13.7500},
    xticklabels={{F1},{Acc},{F1},{Acc},{F1},{Acc},{F1},{Acc},{F1},{Acc},{F1},{Acc}},
    x tick label style={font=\footnotesize, yshift=1pt},
    xtick pos=bottom,
    ymajorgrids,
    grid style={black!12},
    legend columns=4,
    legend image code/.code={
        \draw[#1, draw=black, line width=0.25pt]
            (0cm,-0.10cm) rectangle (0.26cm,0.10cm);
    },
    legend style={
        at={(0.5,1.0)},
        anchor=south,
        draw=none,
        font=\footnotesize,
        /tikz/every even column/.append style={column sep=0.35cm}
    },
    enlarge x limits=0.05,
    tick align=outside,
    clip=false,
]
\addplot+[ybar, bar width=5.8pt, bar shift=-10.2pt,
    draw=black, line width=0.25pt,
    fill=BaselineBlue!35, pattern=horizontal lines,
    pattern color=BaselineBlue] coordinates {
(0.0000,99.799235) (1.0000,99.751637)
(2.5500,99.800118) (3.5500,99.753417)
(5.1000,99.801082) (6.1000,99.754629)
(7.6500,99.757961) (8.6500,99.708420)
(10.2000,99.783795) (11.2000,99.736453)
(12.7500,99.724238) (13.7500,99.677009)
};

\addplot+[ybar, bar width=5.8pt, bar shift=-3.4pt,
    draw=black, line width=0.25pt, fill=BaselineBlue] coordinates {
(0.0000,87.240667) (1.0000,85.994263)
(2.5500,87.343378) (3.5500,86.130639)
(5.1000,85.850145) (6.1000,84.697148)
(7.6500,98.540107) (8.6500,98.214801)
(10.2000,99.744290) (11.2000,99.688369)
(12.7500,99.678852) (13.7500,99.624022)
};

\addplot+[ybar, bar width=5.8pt, bar shift=2.8pt,
    draw=black, line width=0.25pt, fill=RiskTeal] coordinates {
(0.0000,68.998256) (1.0000,70.708754)
(2.5500,69.255238) (3.5500,70.970589)
(5.1000,64.184137) (6.1000,67.446941)
(7.6500,99.318045) (8.6500,99.183226)
(10.2000,99.354693) (11.2000,99.218921)
(12.7500,99.280290) (13.7500,99.162600)
};

\addplot+[ybar, bar width=5.8pt, bar shift=9.0pt,
    draw=black, line width=0.25pt, fill=EfficiencyOrange] coordinates {
(0.0000,99.516653) (1.0000,99.401511)
(2.5500,99.512076) (3.5500,99.398082)
(5.1000,99.624874) (6.1000,99.536925)
(7.6500,96.866093) (8.6500,96.101469)
(10.2000,99.692839) (11.2000,99.625080)
(12.7500,99.685340) (13.7500,99.631248)
};

\legend{{Transfer Learning},{Zero-shot},{Risk-Averse},{Efficiency-Oriented}}

\node[anchor=north, font=\footnotesize, align=center]
    at (axis cs:0.5000,54.0) {FAS shift};
\node[anchor=north, font=\footnotesize, align=center]
    at (axis cs:3.0500,54.0) {FA shift};
\node[anchor=north, font=\footnotesize, align=center]
    at (axis cs:5.6000,54.0) {A shift};
\node[anchor=north, font=\footnotesize, align=center]
    at (axis cs:8.1500,54.0) {F shift};
\node[anchor=north, font=\footnotesize, align=center]
    at (axis cs:10.7000,54.0) {S shift (WC-MCF)};
\node[anchor=north, font=\footnotesize, align=center]
    at (axis cs:13.2500,54.0) {S shift (Bu-SMF)};
\end{axis}
\end{tikzpicture}
        \caption{Feasibility-decision performance.}
        \label{fig:policy-classification-zs-tl}
    \end{subfigure}

    \vspace{0.5cm}

    \begin{subfigure}{\textwidth}
        \centering
        % Figure 4(b): test-set means over ten partitions per scenario,
% then averaged across scenarios within each shift family.
% FPR and FNR are conditional error rates, expressed as percentages.
% Zero-valued bars are placed at the logarithmic display floor.
% Requires qot_sdm_plot_preamble.tex in the manuscript preamble.
\begin{tikzpicture}
\begin{axis}[
    width=0.98\textwidth,
    height=0.245\textheight,
    ybar,
    ymode=log,
    log origin=infty,
    ymin=0.005,
    ymax=180,
    ylabel={Rate (\%)},
    ylabel style={
        at={(axis description cs:-0.055,0.5)},
        anchor=south
    },
    xtick={0,1,2,3,4,5},
    xticklabels={{FAS shift},{FA shift},{A shift},{F shift},{S shift (WC-MCF)},{S shift (Bu-SMF)}},
    x tick label style={font=\footnotesize, align=center},
    yticklabel style={font=\footnotesize},
    xtick pos=bottom,
    ytick pos=left,
    ytick={0.01,0.1,1,10,100},
    yticklabels={$10^{-2}$,$10^{-1}$,$10^{0}$,$10^{1}$,$10^{2}$},
    extra y ticks={0.005},
    extra y tick labels={0},
    ymajorgrids,
    grid style={black!12},
    legend columns=2,
    legend image code/.code={
        \draw[#1, draw=black, line width=0.25pt]
            (0cm,-0.10cm) rectangle (0.26cm,0.10cm);
    },
    legend style={
        at={(1.005,0.99)},
        anchor=north east,
        draw=none,
        fill=none,
        font=\scriptsize,
        /tikz/every even column/.append style={column sep=0.10cm}
    },
    xmin=-0.55,
    xmax=5.55,
    enlarge x limits=false,
    tick align=outside,
    clip=false,
    nodes near coords,
    point meta=explicit symbolic,
    every node near coord/.append style={
        font=\scriptsize,
        text=black,
        rotate=90,
        anchor=west,
        xshift=-1pt,
        yshift=0pt
    },
]

\addplot+[ybar, bar width=5.0pt, bar shift=-15.0pt,
    draw=black, line width=0.25pt, fill=BaselineBlue]
coordinates {
(0,0.005) [0.0]
(1,0.005) [0.0]
(2,0.005) [0.0]
(3,4.490960) [4.5]
(4,0.294477) [0.3]
(5,0.322068) [0.3]
};

\addplot+[ybar, bar width=5.0pt, bar shift=-9.0pt,
    draw=black, line width=0.25pt, fill=BaselineBlue!55]
coordinates {
(0,22.631055) [22.6]
(1,22.469343) [22.5]
(2,24.791675) [24.8]
(3,0.005) [0.0]
(4,0.322194) [0.3]
(5,0.414072) [0.4]
};

\addplot+[ybar, bar width=5.0pt, bar shift=-3.0pt,
    draw=black, line width=0.25pt, fill=RiskTeal]
coordinates {
(0,0.005) [0.0]
(1,0.005) [0.0]
(2,0.005) [0.0]
(3,0.117937) [0.1]
(4,0.011812) [0.01]
(5,0.016942) [0.02]
};

\addplot+[ybar, bar width=5.0pt, bar shift=3.0pt,
    draw=black, line width=0.25pt, fill=RiskTeal!60]
coordinates {
(0,47.330163) [47.3]
(1,47.030160) [47.0]
(2,52.740932) [52.7]
(3,1.277855) [1.3]
(4,1.274463) [1.3]
(5,1.417216) [1.4]
};

\addplot+[ybar, bar width=5.0pt, bar shift=9.0pt,
    draw=black, line width=0.25pt, fill=EfficiencyOrange]
coordinates {
(0,0.869583) [0.9]
(1,0.783253) [0.8]
(2,0.641856) [0.6]
(3,9.807494) [9.8]
(4,0.238842) [0.2]
(5,0.453256) [0.5]
};

\addplot+[ybar, bar width=5.0pt, bar shift=15.0pt,
    draw=black, line width=0.25pt, fill=EfficiencyOrange!60]
coordinates {
(0,0.432251) [0.4]
(1,0.490053) [0.5]
(2,0.350979) [0.4]
(3,0.005) [0.0]
(4,0.456963) [0.5]
(5,0.309100) [0.3]
};

\legend{{FPR - Zero-shot},{FNR - Zero-shot},
        {FPR - Risk-Averse},{FNR - Risk-Averse},
        {FPR - Efficiency-Oriented},{FNR - Efficiency-Oriented}}

\end{axis}
\end{tikzpicture}
        \caption{\Acf{FPR} and \acf{FNR} across shift families.}
        \label{fig:policy_error_rate}
    \end{subfigure}

    \caption{\replaced{Feasibility-decision performance by source--target shift family. Values are means over ten target-domain calibration-test partitions per scenario and then across scenarios in each family. (a) F1 score and accuracy for \ac{TL}, \ac{ZS}, and the \ac{CRC}-based \ac{RA} and \ac{EO} policies. (b) \ac{FPR} and \ac{FNR} for \ac{ZS}, \ac{RA}, and \ac{EO}.}{Performance of the proposed \ac{CRC}-based feasibility policies across source--target shift families. (a) Feasibility-classification performance for the \ac{TL}, \ac{ZS}, \ac{RA}, and \ac{EO} approaches. (b) Corresponding \ac{FPR} and \ac{FNR} values.}}
\label{fig:policy-results}
\end{figure*}
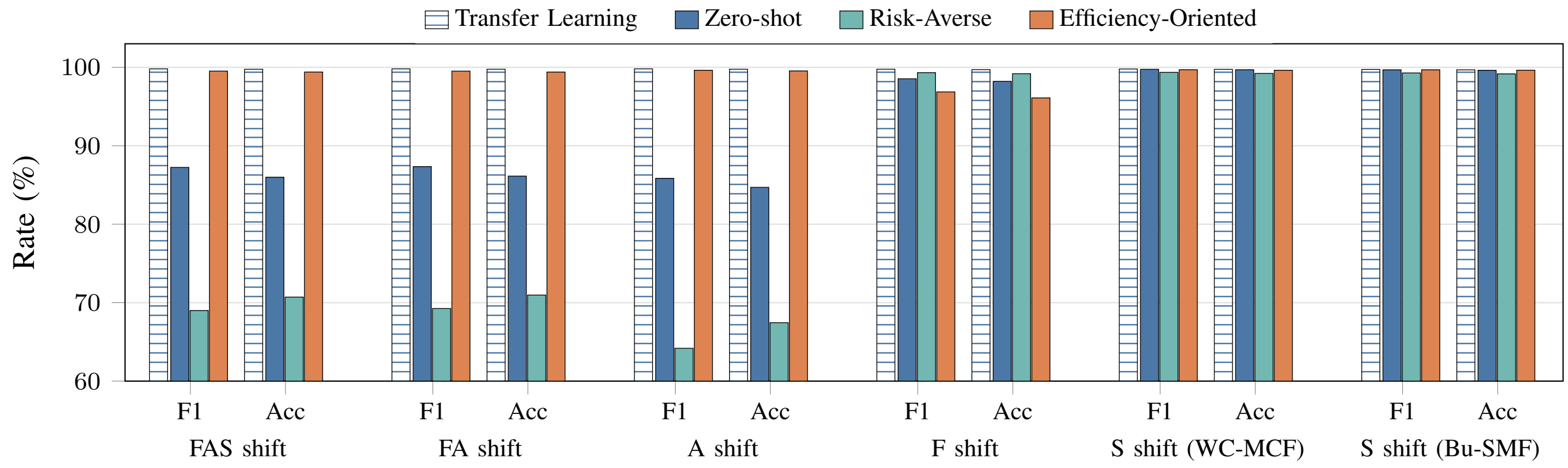
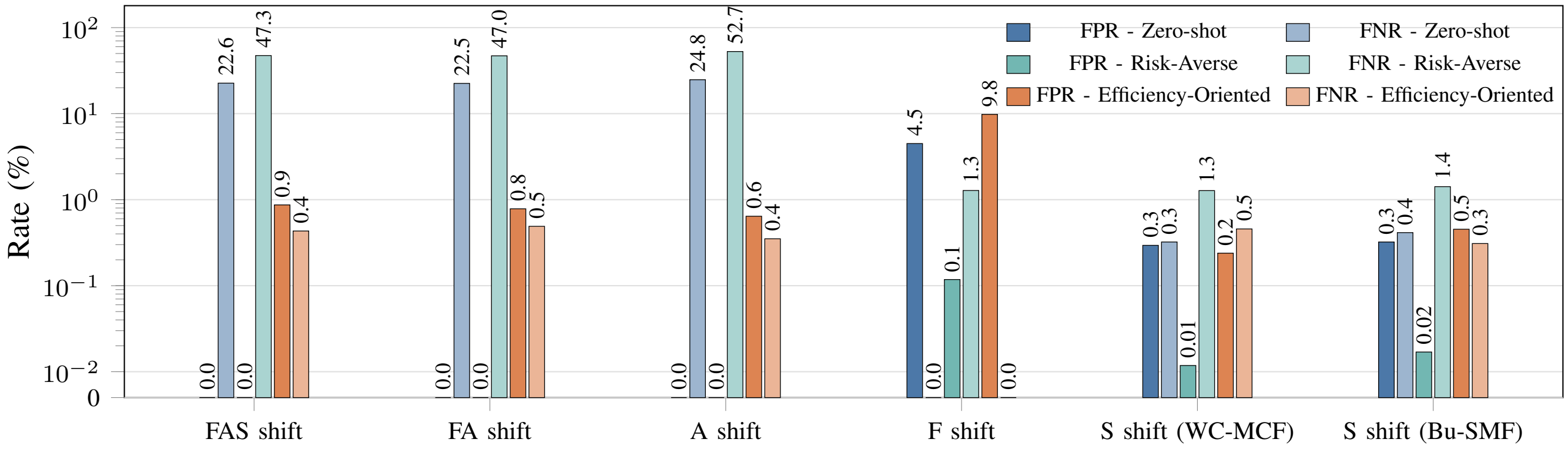

\subsection{Source--Target Shift Characterization}
\label{subsec_shift_characterization}

Fig.~\ref{fig:shift_feature_heatmap} reports the average feature shift contribution score $C_j$ for each source--target domain shift family, where each family consists of multiple shift scenarios.
\replaced{The largest average contribution scores are observed for \ac{FAS} and A shift, indicating that amplifier-related changes play a major role in the source--target mismatch, especially when combined with fiber and switching changes.}{The largest average contribution scores are observed for \ac{FAS} and A shift, indicating that amplifier-related changes play a major role in the source–-target mismatch, especially when combined with fiber and switching changes.}
Across feature groups, the strongest shift is concentrated in link occupation and spectrum-related features.
In particular, average link occupation (LO-Avg), carrier frequency (CF), and active channels (AC) show the most visible contributions in the strongest domain shift families. This suggests that changes in technology have a considerable impact on the distribution and relationships among features related to link occupation and spectrum allocation.

The \ac{FA} shift family also incur feature shifts from link occupation features, although its overall shift intensity is lower than that of \ac{FAS} and A shift.
This is consistent with the fact that amplifier changes affect the resulting \ac{OSNR} behavior and can alter the relation between network load (co-propagating channels) and \ac{QoT}.
By comparison, F shift falls in the lowest bin for every feature, and the S shift (WC-MCF) and S shift (Bu-SMF) families are only marginally above it, suggesting that changing the fiber or the switching architecture alone perturbs the \ac{LP} feature distributions, and their relation to \ac{QoT}, far less than amplifier-related or mixed-technology changes do.
The differences observed across the shift families allow us to study the framework under a broad range of source--target mismatches, from switching-only changes to combined fiber, amplifier, and switching transitions.
The F shift family, however, shows a negligible $C_j$, yet the transition is not harmless, as it yields a wider band than either S shift family (Sec.~\ref{subsec:coverage}).
A fiber-only change leaves the feature distributions essentially intact while altering their relation to \ac{QoT}, and the product in~\eqref{eq:shift_contribution} vanishes in exactly that case, so $C_j$ must be read and analyzed in combination with $\lambda^\star$.

\subsection{LTT Uncertainty Bands Coverage}
\label{subsec:coverage}

\replaced{Table~\ref{tab:ltt-coverage-family-zs} compares the target-domain test coverage and full width of the \ac{LTT} with a source-only \ac{QR} uncertainty bands for each shift family. Source-only \ac{QR} assesses whether intervals learned in the source domain retain their nominal coverage after a technology shift. Unlike \ac{LTT}, it does not receive any target-domain calibration samples. Across the 170 scenario--partition trials, \ac{LTT} achieves at least 90\% coverage in every trial (minimum: 90.48\%; mean: 91.81\%). On held-out source-domain validation data, the nominal 90\% \ac{QR} band achieves 89.64\% mean coverage (89.44--89.85\% across seven source domains). On target-domain test data, however, it does not reach 90\% coverage in any of the trials and has a mean coverage of 29.70\%. Its mean full width is 0.209~dB, compared with 2.244~dB for \ac{LTT}, but these narrow bands do not capture the shift-induced prediction errors. We exclude \ac{QR} from the decision-policy comparison because its target-domain test coverage remains below the 90\% target ($1-\alpha$, with $\alpha=0.10$) in all 170 trials. Scenario-level results for all 17 transitions are reported in Supplementary Table~S3.}{Table~\ref{tab:ltt-coverage-family-zs} reports, per shift family, the calibration and test coverage, i.e., the fraction of calibration and target-domain test \acp{LP} whose true \ac{OSNR} falls inside the selected band, together with $\lambda^\star$.
The two coverages are closely aligned for every family and remain above the nominal 90\% target induced by $\alpha=0.10$ at $1-\delta=0.95$, showing that the procedure generalizes to unseen target-domain \acp{LP}.}

\replaced{This difference between \ac{LTT} and \ac{QR} uncertainty bands is also visible in Fig.~\ref{fig:bands-selected-shifts-zs}. For \ac{LTT} bands, Amplifier-related and mixed shifts require full widths of 4.2--4.7~dB, whereas switching-only shifts require about 0.2~dB. Wider bands reflect larger target-domain residuals, although their effect on feasibility decisions also depends on the residual direction and the \ac{OSNR} margin to the threshold. In particular, a wide \ac{LTT} band can cause the \ac{RA} policy to reject otherwise feasible \acp{LP}, reducing resource utilization. Narrower bands would require smaller residuals or less finite-sample slack, for example by improving or adapting the predictor using data disjoint from calibration and then recalibrating it, or by collecting more representative calibration data. Simply reducing the band width at fixed $\alpha$, $\delta$, and residuals would invalidate the coverage guarantee. The narrow \ac{QR} bands remain close to the source-domain estimates but often miss the target-domain \ac{OSNR}, particularly for amplifier-related shifts. In contrast, \ac{CRC} widens the band according to the shifted residuals. Thus, interval width should be interpreted jointly with coverage, rather than as evidence of reliable uncertainty on its own.}{The selected widths $\lambda^\star$ reflect the uncertainty of the \ac{OSNR} predictions.
Amplifier-related and mixed-technology shifts require the widest bands, the fiber-only shift an intermediate one, and the switching-only shifts the narrowest, by an order of magnitude.
Fig.~\ref{fig:bands-selected-shifts-zs} shows the same effect per \ac{LP}.
The true \ac{OSNR} lies inside the band for the \acp{LP} shown, the per-\ac{LP} counterpart of the aggregate coverage, while the width tracks the shift severity, visibly wider for \ac{FAS} and A shift than for F shift.
The framework thus exposes the cost of reusing the source-trained regressor as an explicit \ac{OSNR} margin rather than leaving it hidden in the point-prediction residuals.
Crucially, $\lambda^\star$ also captures aspects that complement the $C_j$ analysis.
The fiber-only shift, whose $C_j$ is negligible, still yields a markedly wider band than the switching-only shifts, so the width measures what the mismatch costs the deployed regressor whether or not the feature distributions move.}

\added{We examined the sensitivity to calibration set size by varying it from 500 samples to the full sets of 165,511--180,629 samples. With 500 samples, \ac{LTT} selected a band in only 107/170 trials; therefore, we do not report aggregate performance. From 1,000 samples onward, a band was selected in all trials and attained at least 90\% test coverage. Increasing the calibration size reduced overcoverage and band width in every shift family, while amplifier-related and mixed shifts remained the widest. Thus, 1,000 samples was the smallest evaluated calibration set size satisfying both conditions, but this value cannot be generalized to other scenarios not investigated in this paper. Details are provided in Supplementary Table~S3.}

\subsection{Feasibility-Decision Performance}
\label{subsec:feasibility_decision}

\deleted{Fig. \ref{fig:policy-classification-zs-tl} reports F1-score and accuracy, while Fig. \ref{fig:policy_error_rate} reports the corresponding \ac{FPR} and \ac{FNR}. As described in Section~\ref{sec:exp_setup}, \ac{ZS} and \ac{TL} make decisions from point \ac{OSNR} predictions, whereas \ac{RA} and \ac{EO} use the \ac{LTT} uncertainty band. The \ac{TL} policy serves as a practical upper reference, and all methods are evaluated on the same held-out test subset.}

\deleted{Under stronger shift families, the \ac{ZS} baseline falls short of the proposed uncertainty-aware policies.
Under \ac{FAS}, \ac{FA}, and A shift, the \ac{ZS} policy attains F1-scores of $87.2\%$, $87.4\%$, and $85.9\%$, respectively. In contrast, , the \ac{EO} policy achieves an F1-score above $99.5\%$ for these shift families, closely approaching the \ac{TL} policy without requiring fine-tuning of the source-trained regressor. For the S shift families, where the source--target mismatch is weaker, all methods achieve high performance, and the \ac{ZS} policy already reaches F1-scores close to $99.8\%$. Operationally, this means that when the technology change has limited impact on the feasibility boundary, the proposed policies preserve the already accurate decisions of the source-trained regressor. The F shift case shows a slightly different behavior, with \ac{RA} policy achieving the highest F1-score of $99.3\%$, compared with $98.5\%$ for the \ac{ZS} policy and slightly lower performance for \ac{EO} policy.}

\deleted{Fig.\ref{fig:policy_error_rate} explains these classification results through the two error types. For \ac{FAS}, \ac{FA}, and A shift, the performance degradation of the \ac{ZS} policy is mainly caused by high \ac{FNR}, exceeding $22\%$ in all three cases. This means that a considerable number of feasible \acp{LP} are incorrectly rejected. The \ac{EO} policy reduces this error to below $0.5\%$, thereby preserving more feasible provisioning opportunities. By contrast, the \ac{RA} policy nearly eliminates false positives across all shift families, with \ac{FPR} close to zero, but at the cost of higher \ac{FNR}.}
\deleted{In the F shift case, the dominant \ac{ZS} policy error is instead \ac{FPR}; consequently, the \ac{RA} policy attains the best performance in this case by reducing the \ac{FPR} to 0.1\%, whereas the \ac{EO} policy keeps false negatives low but increases the \ac{FPR} to 9.8\%.}

\deleted{These findings show that the best policy depends on the dominant error type within each domain shift family. When the point-prediction baseline policy mainly rejects feasible \acp{LP}, as in \ac{FAS}, \ac{FA}, and A shift, the \ac{EO} policy is preferable because it reduces the number of missed feasible \acp{LP}.
When the dominant error is accepting infeasible \acp{LP}, as in the F shift case, \ac{RA} is preferable because it suppresses false positives.}
\deleted{From an operator perspective, the \ac{EO} policy preserves provisioning opportunities by correctly labeling most feasible \acp{LP} that the \ac{ZS} policy would reject, making it suitable when the priority is to reduce overly conservative rejection. Conversely, \ac{RA} prioritizes avoiding the acceptance of infeasible \ac{LP}, which is important when failed provisioning or service degradation is costly.
These results show that the proposed framework supports multiple operating modes according to the operator's objective, rather than enforcing a single fixed decision rule.}

\added{Figs.~\ref{fig:policy-classification-zs-tl} and~\ref{fig:policy_error_rate} show decision performance and error type for the \ac{ZS}, \ac{TL}, and proposed \ac{CRC}-based policies. Under \ac{FAS}, \ac{FA}, and A shifts, \ac{EO} raises the degraded \ac{ZS} accuracy from about 85\% to above 99\%, appraching \ac{TL} levels. Switching-only shifts leave all methods near 99\%. F shift reverses the ranking: \ac{RA} is the strongest non-retraining method and \ac{EO} falls below \ac{ZS}. Across the 170 scenario--partition trials, the mean \ac{RA} and \ac{EO} accuracies are 85.26\% and 98.92\%, respectively. Using only 1,000 instead of all calibration samples drops them by less than 1\%. Within each shift family, the change is at most 1.25 percentage points (Supplementary Table~S4).}

\added{The results show that residual direction depends on the technology shift: amplifier-related shifts cause underestimation and high \ac{FNR}, which \ac{EO} reduces, whereas F shift causes overestimation and high \ac{FPR}, which \ac{RA} reduces. The direction of the shift cannot be assumed before target calibration data are observed. The symmetric \ac{LTT} uncertainty band is therefore shift-agnostic and attains the target coverage in all trials.
Once the direction is observed, however, equal widths may place unnecessary width on the less relevant side of the band. An asymmetric band could use different lower and upper widths to address this, but we leave its calibration and evaluation outside the scope of this study. %An asymmetric extension, $[\hat f(x)-\lambda_{-},\hat f(x)+\lambda_{+}]$, could calibrate both tails separately, with $\alpha_{-}+\alpha_{+}=\alpha$, assigning a smaller risk to the lower bound for \ac{RA} or the upper bound for \ac{EO}.
}

\added{\ac{TL} is an upper reference: it uses the same target-data budget but requires model access and retraining, whereas \ac{CRC} recalibrates a fixed predictor; the two can also be combined}~\cite{kiarash2026ofc}\added{. The guarantee requires calibration--test exchangeability. Noise or aging represented during calibration can widen the band, whereas continued drift, abrupt instability, or evolving noise conditions are not covered. %Deployment should use rolling calibration, residual monitoring, and recalibration; aging/noise traces and adaptive conformal updates are future work.
}

\section{Conclusion}
\label{sec:conclusion}

\replaced{Technology upgrades can make a source-trained \ac{QoT} regressor unreliable without revealing the loss in its point predictions. The proposed post-hoc \ac{CRC} framework addresses this without retraining: target calibration data construct \ac{LTT} uncertainty bands with finite-sample control of target miscoverage, feature shift scores characterize mismatch, and \ac{RA} and \ac{EO} use the bands for feasibility decisions. The guarantee applies to coverage; policy performance is evaluated empirically.}{This paper proposed a post-hoc \ac{CRC}-based decision-support framework for \ac{QoT}-aware \ac{LP} feasibility assessment under source-–target mismatch. The framework keeps a source-trained \ac{QoT} regressor fixed, converts its point \ac{QoT} predictions into \ac{LTT} uncertainty bands whose width is backed by an explicit, finite-sample bound on the miscoverage risk, and uses these bands to support uncertainty-aware feasibility decisions through \ac{RA} and \ac{EO} policies. A feature shift contribution score was used alongside it to characterize how transitions in fiber, amplifier, and switching technologies affect the relationship between \ac{LP} features and \ac{QoT} values.}

\replaced{Across 17 shifts and ten target-data partitions, \ac{CRC} met the 90\% test-coverage target in all 170 trials, whereas source-only \ac{QR} met it in none. With 1,000 samples, \ac{LTT} selected a band in every trial with decision performance close to the full-calibration result; with 500 samples, it selected a band in 107/170 trials. Wider amplifier-related bands expose uncertainty from reusing the source regressor, while \ac{RA} and \ac{EO} outcomes show the trade-off between safety and resource utilization.}
{Numerical experiments on open \ac{SDM} \ac{QoT} datasets over the CONUS topology, spanning 17 source--target scenarios, showed that the framework maintains above-target empirical coverage across all evaluated shift families and, under the most severe technology-driven shifts, improves feasibility-decision accuracy from about 85\% to above 99\%, approaching a \ac{TL} policy without retraining the underlying \ac{QoT} regressor.}

\replaced{The guarantee requires calibration--test exchangeability. Aging, evolving noise, sensor shift, or link instability not represented during calibration require monitoring and recalibration. Future work should consider these effects, in addition to rolling calibration, and asymmetric bands. The same approach may support threshold-based regression tasks in optical networks.}{These results highlight the potential of \ac{CRC} to support trustworthy AI/ML-assisted optical network operation and automation by explicitly quantifying prediction uncertainty and incorporating it into operational decision-making under deployment shift. Since the \ac{CRC} and policy layers only consume the residuals of a fixed predictor, the same principle extends to other regression-driven optical network tasks in which a decision threshold is applied to an uncertain estimate, such as margin assignment or failure-risk assessment.}

\ifCLASSOPTIONcaptionsoff
  \newpage
\fi
%\IEEEtriggeratref{36}
%\bibliographystyle{IEEEtrandoi}
\bibliographystyle{IEEEtran}
\bibliography{bibtex/IEEEabrv, references}


\end{document}

\maketitle
\acresetall

This supplement provides the scenario-level results underlying the repeated-partition, source-only \ac{QR}, and calibration-size analyses reported in the main manuscript. It contains no additional claims beyond those in the manuscript. It documents the experimental protocol and corresponding numerical results in greater detail.

\section{Repeated-Partition Protocol}
\label{sec:supp-protocol}

This section details the repeated-partition and \ac{LTT} uncertainy band-selection procedures described in Sec.~IV of the main manuscript.
These procedures underlie the coverage and the corresponding interval-width results in Table~II and the feasibility-decision results in Fig.~4.

The source-domain training and validation sets and the target-domain adaptation set were kept fixed. For each source--target scenario, the union of the original target calibration and test sets was reshuffled with ten seeds: 101, 211, 307, 401, 503, 601, 701, 809, 907, and 1009. We retained the original calibration and test sizes, and the two subsets were disjoint within each partition. Because different partitions draw from the same target-domain pool, their overlap is possible. Consequently, the reported standard deviations quantify sensitivity to the calibration--test partition and are not standard errors over independent datasets. Within each partition, we evaluated \ac{CRC} and \ac{QR} on exactly the same test samples.

The \ac{CRC} experiments used $\alpha=0.10$ and $\delta=0.05$. For each calibration set, 60 candidate values of $\lambda$ were constructed from its absolute source-model residuals, and \ac{LTT} selected the smallest candidate satisfying the stated risk criterion. If no candidate satisfied the criterion, no band was selected and that trial was recorded as such rather than assigned a width or coverage value.

For the source-only \ac{QR} baseline, two \ac{XGB} quantile regressors estimated $\widehat q_{0.05}(x)$ and $\widehat q_{0.95}(x)$. Each endpoint was averaged over five models trained with seeds 42, 53, 64, 75, and 86. The models used only source-domain data and received neither target-domain training nor target-domain calibration.

Table~\ref{tab:supp-sample-counts} reports the target-domain sample counts used in the full-calibration experiments. These counts are preserved for every partition of a given scenario.

\begin{table}[H]
\centering
\caption{Target-domain calibration and test sample counts for the 17 source--target scenarios.}
\label{tab:supp-sample-counts}
%\footnotesize
\small
\renewcommand{\arraystretch}{1.02}
\setlength{\tabcolsep}{10pt}

\begin{tabular}{clrr}
\toprule
\textbf{Shift family}
& \textbf{Source$\rightarrow$Target}
& \textbf{Calibration}
& \textbf{Test} \\
\midrule

\multirow{2}{*}{\ac{FAS} shift}
& $3\rightarrow7$ & 175,568 & 526,704 \\
& $3\rightarrow8$ & 179,579 & 538,738 \\
\midrule

\multirow{3}{*}{\ac{FA} shift}
& $1\rightarrow7$ & 175,568 & 526,704 \\
& $2\rightarrow8$ & 179,579 & 538,738 \\
& $3\rightarrow9$ & 180,629 & 541,887 \\
\midrule

\multirow{3}{*}{A shift}
& $4\rightarrow7$ & 175,568 & 526,704 \\
& $5\rightarrow8$ & 179,579 & 538,738 \\
& $6\rightarrow9$ & 180,629 & 541,887 \\
\midrule

\multirow{3}{*}{F shift}
& $1\rightarrow4$ & 165,558 & 496,674 \\
& $2\rightarrow5$ & 166,279 & 498,837 \\
& $3\rightarrow6$ & 165,511 & 496,534 \\
\midrule

\multirow{4}{*}{S shift (\ac{WC-MCF})}
& $6\rightarrow4$ & 165,558 & 496,674 \\
& $6\rightarrow5$ & 166,279 & 498,837 \\
& $9\rightarrow7$ & 175,568 & 526,704 \\
& $9\rightarrow8$ & 179,579 & 538,738 \\
\midrule

\multirow{2}{*}{S shift (\ac{Bu-SMF})}
& $3\rightarrow1$ & 172,055 & 516,166 \\
& $3\rightarrow2$ & 172,821 & 518,463 \\

\bottomrule
\end{tabular}
\end{table}

\clearpage
\section{Scenario-Level CRC and QR Results}
\label{sec:supp-crc-qr}

%Table~\ref{tab:supp-crc-qr-scenarios} gives the test coverage and full interval width for every scenario. Each entry is the mean $\pm$ standard deviation over the ten target-data partitions. The \ac{CRC} band reaches at least 90\% mean test coverage in every scenario. The nominal 90\% source-only \ac{QR} interval falls below 90\% in every scenario, with the largest shortfall under shifts involving amplifier changes.
The manuscript compares the \ac{LTT} uncertainty bands and source-only \ac{QR} by shift family in its Results and Discussion section (Sec.~V, specifically Sec.~V-B and Table~II). Table~\ref{tab:supp-crc-qr-scenarios} gives the corresponding coverage and full interval width for each source--target scenario, as mean $\pm$ standard deviation over ten target-data partitions. The \ac{LTT} uncertainty bands reach at least 90\% mean test coverage in every scenario, whereas source-only \ac{QR} falls below 90\% in every scenario, most markedly under amplifier-related shifts.

\begin{table}[H]
\centering
\caption{Scenario-level test coverage and full interval width for \ac{CRC} and source-only \ac{QR}. Values are the mean $\pm$ standard deviation over ten target-data partitions.}
\label{tab:supp-crc-qr-scenarios}
%\scriptsize
\small
\renewcommand{\arraystretch}{1.02}
\setlength{\tabcolsep}{5pt}

\begin{tabular}{clcccc}
\toprule
& & \multicolumn{2}{c}{\textbf{\ac{LTT} uncertainty band}}
& \multicolumn{2}{c}{\textbf{Source-only \ac{QR}}} \\
\cmidrule(lr){3-4}\cmidrule(lr){5-6}
\textbf{Shift Family}
& \textbf{Source$\rightarrow$Target}
& \textbf{Coverage (\%)}
& \textbf{Width (dB)}
& \textbf{Coverage (\%)}
& \textbf{Width (dB)} \\
\midrule

\multirow{2}{*}{\ac{FAS} shift}
& $3\rightarrow7$ & $91.73\pm0.07$ & $4.220\pm0.001$ & $0.00\pm0.00$ & $0.219\pm0.000$ \\
& $3\rightarrow8$ & $91.57\pm0.10$ & $4.181\pm0.001$ & $0.00\pm0.00$ & $0.219\pm0.000$ \\
\midrule

\multirow{3}{*}{\ac{FA} shift}
& $1\rightarrow7$ & $91.92\pm0.06$ & $4.180\pm0.000$ & $0.00\pm0.00$ & $0.219\pm0.000$ \\
& $2\rightarrow8$ & $91.36\pm0.07$ & $4.153\pm0.000$ & $0.00\pm0.00$ & $0.224\pm0.000$ \\
& $3\rightarrow9$ & $91.21\pm0.13$ & $4.150\pm0.001$ & $0.00\pm0.00$ & $0.219\pm0.000$ \\
\midrule

\multirow{3}{*}{A shift}
& $4\rightarrow7$ & $91.12\pm0.06$ & $4.630\pm0.000$ & $0.00\pm0.00$ & $0.219\pm0.000$ \\
& $5\rightarrow8$ & $93.72\pm4.02$ & $4.649\pm0.053$ & $0.00\pm0.00$ & $0.217\pm0.000$ \\
& $6\rightarrow9$ & $98.46\pm0.02$ & $4.709\pm0.001$ & $0.00\pm0.00$ & $0.212\pm0.000$ \\
\midrule

\multirow{3}{*}{F shift}
& $1\rightarrow4$ & $91.33\pm0.19$ & $0.672\pm0.002$ & $2.80\pm0.01$ & $0.212\pm0.000$ \\
& $2\rightarrow5$ & $91.28\pm0.06$ & $0.702\pm0.001$ & $3.69\pm0.01$ & $0.217\pm0.000$ \\
& $3\rightarrow6$ & $90.81\pm0.06$ & $0.723\pm0.001$ & $3.34\pm0.01$ & $0.217\pm0.000$ \\
\midrule

\multirow{4}{*}{S shift (\ac{WC-MCF})}
& $6\rightarrow4$ & $91.00\pm0.06$ & $0.237\pm0.000$ & $79.34\pm0.03$ & $0.205\pm0.000$ \\
& $6\rightarrow5$ & $91.06\pm0.09$ & $0.208\pm0.000$ & $85.65\pm0.01$ & $0.203\pm0.000$ \\
& $9\rightarrow7$ & $90.93\pm0.08$ & $0.144\pm0.000$ & $79.74\pm0.03$ & $0.161\pm0.000$ \\
& $9\rightarrow8$ & $91.15\pm0.23$ & $0.124\pm0.001$ & $86.07\pm0.02$ & $0.160\pm0.000$ \\
\midrule

\multirow{2}{*}{S shift (\ac{Bu-SMF})}
& $3\rightarrow1$ & $90.97\pm0.06$ & $0.250\pm0.000$ & $78.82\pm0.02$ & $0.215\pm0.000$ \\
& $3\rightarrow2$ & $91.13\pm0.07$ & $0.220\pm0.000$ & $85.49\pm0.02$ & $0.214\pm0.000$ \\

\bottomrule
\end{tabular}
\end{table}

\section{Calibration-Size Sensitivity}
\label{sec:supp-calibration-size}

In this section, we examine how calibration set size affects \ac{LTT} band selection, test coverage, and band width. These results are summarized in the manuscript's Results and Discussion section (Sec.~V, specifically Sec.~V-B).

For each scenario and partition, calibration was repeated using nested subsets of 500, 1,000, 2,500, 5,000, 10,000, 25,000, 50,000, and 100,000 samples, followed by the full calibration set. A separate candidate grid was constructed from each subset; the full-set grid was not reused for the smaller budgets. Table~\ref{tab:supp-calibration-size} reports the test coverage and full band width by shift family.

At 500 samples, \ac{LTT} selected a band in 107 of the 170 scenario--partition trials: 20/20 \ac{FAS}, 30/30 \ac{FA}, 30/30 A, 19/30 F, 5/40 S (WC-MCF), and 3/20 S (Bu-SMF) trials. The 500-sample entries therefore average only the trials in which a band was selected. From 1,000 samples onward, a band was selected in all 170 trials and all selected bands reached at least 90\% test coverage. Thus, 1,000 samples is the smallest evaluated budget with this outcome in every trial; it is an empirical result for these data and is not a universal minimum.

\begin{table}[H]
\centering
\caption{Calibration-size sensitivity by shift family. Each entry gives mean test coverage (\%)/mean full \ac{LTT} uncertainty band width (dB) over the trials in which a band was selected.}
\label{tab:supp-calibration-size}
%\scriptsize
\small
\renewcommand{\arraystretch}{1.05}
\setlength{\tabcolsep}{2.5pt}
\begin{tabular*}{\textwidth}{@{\extracolsep{\fill}}lccccccccc@{}}
\toprule
& \multicolumn{9}{c}{\textbf{Calibration samples}} \\
\cmidrule(lr){2-10}
\textbf{Shift family}
& \textbf{500}
& \textbf{1k}
& \textbf{2.5k}
& \textbf{5k}
& \textbf{10k}
& \textbf{25k}
& \textbf{50k}
& \textbf{100k}
& \textbf{Full} \\
\midrule
\ac{FAS} shift
& 99.58/4.386 & 97.83/4.290 & 97.94/4.292 & 97.99/4.294 & 97.98/4.293
& 91.69/4.200 & 91.63/4.200 & 91.65/4.200 & 91.65/4.200 \\
\ac{FA} shift
& 99.66/4.345 & 98.06/4.254 & 98.06/4.254 & 98.09/4.254 & 97.87/4.251
& 94.93/4.207 & 91.49/4.161 & 91.50/4.161 & 91.49/4.161 \\
A shift
& 99.83/4.822 & 98.42/4.719 & 98.42/4.718 & 98.43/4.719 & 98.40/4.718
& 98.39/4.718 & 96.24/4.687 & 95.74/4.680 & 94.43/4.663 \\
F shift
& 99.29/1.086 & 96.82/0.941 & 94.45/0.830 & 93.21/0.775 & 92.52/0.743
& 91.89/0.721 & 91.40/0.707 & 91.17/0.700 & 91.14/0.699 \\
S shift (WC-MCF)
& 98.41/0.243 & 96.76/0.229 & 94.48/0.203 & 93.38/0.194 & 92.52/0.188
& 91.80/0.183 & 91.22/0.180 & 91.06/0.179 & 91.03/0.178 \\
S shift (Bu-SMF)
& 99.00/0.374 & 96.93/0.304 & 94.75/0.270 & 93.56/0.257 & 92.63/0.248
& 91.85/0.241 & 91.51/0.238 & 91.06/0.235 & 91.05/0.235 \\
\bottomrule
\end{tabular*}
\vspace{2pt}

{\footnotesize At 500 samples, the band-selection counts are 20/20 (FAS), 30/30 (FA), 30/30 (A), 19/30 (F), 5/40 [S (WC-MCF)], and 3/20 [S (Bu-SMF)]. All 170 trials are included at every budget from 1,000 samples onward.}
\end{table}

With 500 samples, the selected bands are generally wider and their coverage is well above the 90\% target. As the calibration budget increases, coverage approaches the target and the mean width decreases. The rate of this change depends on the shift family: amplifier-related and mixed shifts remain substantially wider than switching-only shifts at every budget.

\section{Effect of Calibration Size on the Decision Policies}
\label{sec:supp-policy-sensitivity}

In this section, we assess how calibration-set size affects the performance of the \ac{RA} and \ac{EO} feasibility policies. These results are summarized in the manuscript's Results and Discussion section (Sec.~V, specifically Sec.~V-C).

Table~\ref{tab:supp-policy-sensitivity} reports the test-set results separately for each shift family. The policy parameters are unchanged from the main manuscript: $\epsilon_{\mathrm{RA}}=0$ and $\epsilon_{\mathrm{EO}}=0.1$~dB. Accuracy and conditional decision-error rates are reported only when \ac{LTT} selected a band in all 170 scenario--partition trials at that calibration budget. Therefore, the 500-sample row contains only the band-selection count for each family; the corresponding policy metrics are omitted because a band was selected in only 107/170 trials overall. From 1,000 samples onward, a band was selected in all 170 trials and all policy metrics are reported. The \ac{FPR} and \ac{FNR} are empirical quantities and are not controlled by the \ac{CRC} coverage guarantee.

\begingroup
\footnotesize
%\footnotesize
\setlength{\LTcapwidth}{\textwidth}
\setlength{\tabcolsep}{5.5pt}
\renewcommand{\arraystretch}{1.02}

\begin{longtable}{clccccc}
\caption{Decision-policy sensitivity to calibration size, separated by shift family. Results are test-set means over the scenario--partition trials within each family. Accuracy, \ac{FPR}, and \ac{FNR} are in percent. A dash indicates that policy metrics are not reported because a band was not selected in all 170 trials at that budget.}
\label{tab:supp-policy-sensitivity}\\
\toprule
\textbf{Shift family}
& \textbf{Cal. samples}
& \textbf{Bands selected}
& \textbf{RA accuracy}
& \textbf{RA FPR}
& \textbf{EO accuracy}
& \textbf{EO FNR} \\
\midrule
\endfirsthead

\multicolumn{7}{c}{\tablename\ \thetable\ (continued)}\\
\toprule
\textbf{Shift family}
& \textbf{Cal. samples}
& \textbf{Bands selected}
& \textbf{RA accuracy}
& \textbf{RA FPR}
& \textbf{EO accuracy}
& \textbf{EO FNR} \\
\midrule
\endhead

\midrule
\multicolumn{7}{r}{Continued on next page}\\
\endfoot

\bottomrule
\endlastfoot

\multirow{9}{*}{\ac{FAS} shift}
& 500  & 20/20 & --    & --    & --    & --    \\
& 1k   & 20/20 & 70.43 & 0.000 & 99.35 & 0.171 \\
& 2.5k & 20/20 & 70.42 & 0.000 & 99.34 & 0.166 \\
& 5k   & 20/20 & 70.41 & 0.000 & 99.34 & 0.163 \\
& 10k  & 20/20 & 70.41 & 0.000 & 99.34 & 0.164 \\
& 25k  & 20/20 & 70.71 & 0.000 & 99.40 & 0.431 \\
& 50k  & 20/20 & 70.71 & 0.000 & 99.40 & 0.433 \\
& 100k & 20/20 & 70.71 & 0.000 & 99.40 & 0.432 \\
& Full & 20/20 & 70.71 & 0.000 & 99.40 & 0.432 \\
\midrule

\multirow{9}{*}{\ac{FA} shift}
& 500  & 30/30 & --    & --    & --    & --    \\
& 1k   & 30/30 & 70.68 & 0.000 & 99.38 & 0.179 \\
& 2.5k & 30/30 & 70.68 & 0.000 & 99.39 & 0.180 \\
& 5k   & 30/30 & 70.67 & 0.000 & 99.38 & 0.179 \\
& 10k  & 30/30 & 70.69 & 0.000 & 99.39 & 0.188 \\
& 25k  & 30/30 & 70.83 & 0.000 & 99.41 & 0.315 \\
& 50k  & 30/30 & 70.97 & 0.000 & 99.40 & 0.490 \\
& 100k & 30/30 & 70.97 & 0.000 & 99.40 & 0.490 \\
& Full & 30/30 & 70.97 & 0.000 & 99.40 & 0.490 \\
\midrule

\multirow{9}{*}{A shift}
& 500  & 30/30 & --    & --    & --    & --    \\
& 1k   & 30/30 & 67.21 & 0.000 & 99.52 & 0.172 \\
& 2.5k & 30/30 & 67.21 & 0.000 & 99.52 & 0.174 \\
& 5k   & 30/30 & 67.21 & 0.000 & 99.52 & 0.173 \\
& 10k  & 30/30 & 67.21 & 0.000 & 99.52 & 0.175 \\
& 25k  & 30/30 & 67.21 & 0.000 & 99.52 & 0.175 \\
& 50k  & 30/30 & 67.35 & 0.000 & 99.54 & 0.263 \\
& 100k & 30/30 & 67.38 & 0.000 & 99.54 & 0.286 \\
& Full & 30/30 & 67.45 & 0.000 & 99.54 & 0.351 \\
\midrule

\multirow{9}{*}{F shift}
& 500  & 19/30 & --    & --    & --    & --    \\
& 1k   & 30/30 & 98.24 & 0.017 & 94.85 & 0.000 \\
& 2.5k & 30/30 & 98.69 & 0.053 & 95.44 & 0.000 \\
& 5k   & 30/30 & 98.90 & 0.075 & 95.72 & 0.000 \\
& 10k  & 30/30 & 99.02 & 0.093 & 95.88 & 0.000 \\
& 25k  & 30/30 & 99.11 & 0.105 & 95.99 & 0.000 \\
& 50k  & 30/30 & 99.16 & 0.113 & 96.06 & 0.000 \\
& 100k & 30/30 & 99.18 & 0.117 & 96.10 & 0.000 \\
& Full & 30/30 & 99.18 & 0.118 & 96.10 & 0.000 \\
\midrule

\multirow{9}{*}{S shift (WC-MCF)}
& 500  & 5/40  & --    & --    & --    & --    \\
& 1k   & 40/40 & 99.02 & 0.004 & 99.65 & 0.281 \\
& 2.5k & 40/40 & 99.12 & 0.007 & 99.65 & 0.362 \\
& 5k   & 40/40 & 99.16 & 0.008 & 99.64 & 0.396 \\
& 10k  & 40/40 & 99.18 & 0.009 & 99.63 & 0.420 \\
& 25k  & 40/40 & 99.20 & 0.011 & 99.63 & 0.438 \\
& 50k  & 40/40 & 99.21 & 0.012 & 99.63 & 0.452 \\
& 100k & 40/40 & 99.22 & 0.012 & 99.63 & 0.456 \\
& Full & 40/40 & 99.22 & 0.012 & 99.63 & 0.457 \\
\midrule

\multirow{9}{*}{S shift (Bu-SMF)}
& 500  & 3/20  & --    & --    & --    & --    \\
& 1k   & 20/20 & 98.95 & 0.006 & 99.57 & 0.156 \\
& 2.5k & 20/20 & 99.05 & 0.010 & 99.61 & 0.222 \\
& 5k   & 20/20 & 99.09 & 0.012 & 99.62 & 0.253 \\
& 10k  & 20/20 & 99.12 & 0.014 & 99.63 & 0.275 \\
& 25k  & 20/20 & 99.14 & 0.015 & 99.63 & 0.292 \\
& 50k  & 20/20 & 99.15 & 0.016 & 99.63 & 0.300 \\
& 100k & 20/20 & 99.16 & 0.017 & 99.63 & 0.309 \\
& Full & 20/20 & 99.16 & 0.017 & 99.63 & 0.309 \\

\end{longtable}
\endgroup